%% file: main.tex
\documentclass{article} 
\PassOptionsToPackage{sort&compress}{natbib}
\PassOptionsToPackage{hyphens}{url}
\usepackage[T1]{fontenc}
\usepackage{times}
\usepackage[noupquote]{inconsolata}  
\usepackage{iclr2027_conference}
\setcitestyle{numbers,square,comma}

\input{math_commands.tex}

\usepackage{hyperref}
\usepackage{url}
\hypersetup{colorlinks=true, citecolor=blue, urlcolor=blue, linkcolor=red}
\usepackage{amsmath,amssymb}
\usepackage{booktabs}
\usepackage{multirow}
\usepackage{graphicx}
\usepackage{float}
\usepackage{subcaption}
\usepackage{xcolor}
\usepackage{colortbl}
\makeatletter
\g@addto@macro\normalsize{\setlength\abovedisplayskip{7pt}\setlength\belowdisplayskip{7pt}\setlength\abovedisplayshortskip{3pt}\setlength\belowdisplayshortskip{3pt}}
\makeatletter
\def\section{\@startsection {section}{1}{\z@}{-1.2ex plus -0.3ex minus -.2ex}{0.7ex plus 0.2ex minus 0.2ex}{\large\sc\raggedright}}
\def\subsection{\@startsection{subsection}{2}{\z@}{-1.0ex plus -0.3ex minus -.2ex}{0.3ex plus .2ex}{\normalsize\sc\raggedright}}
\def\paragraph{\@startsection{paragraph}{4}{\z@}{0.5ex plus 0.2ex minus .2ex}{-1em}{\normalsize\bf}}
\g@addto@macro\normalsize{\setlength\abovedisplayskip{4pt}\setlength\belowdisplayskip{4pt}\setlength\abovedisplayshortskip{1pt}\setlength\belowdisplayshortskip{1pt}}
\makeatother
\makeatother
\usepackage{pifont}
\usepackage{tikz}
\usetikzlibrary{calc} 

\newcommand{\probejs}{\textsc{Probe-JS}}
\newcommand{\lens}{\textsc{Lens-Shift}}
\newcommand{\midpoint}{\textsc{Midpoint}}
\newcommand{\entropybl}{\textsc{Entropy}}
\definecolor{gold}{HTML}{D4AF37}
\definecolor{silver}{HTML}{8C9196}
\definecolor{bronze}{HTML}{CD7F32}
\definecolor{ablref}{gray}{0.90}
\definecolor{ablpj}{HTML}{DCE8F5}
\definecolor{abllens}{HTML}{FBE6D0}
\DeclareRobustCommand{\mdl}[2]{\,\tikz[baseline=-0.4ex]\node[circle,fill=#1,text=white,inner sep=0.6pt,font=\bfseries\fontsize{4.5}{4.5}\selectfont]{#2};}
\newcommand{\best}[1]{$\mathbf{#1}$\mdl{gold}{1}}
\newcommand{\secondbest}[1]{$#1$\mdl{silver}{2}}

\newcommand{\JS}{\mathrm{JS}}

\newcommand{\at}[1]{{\scriptsize\textcolor{gray}{@$#1$}}}
\newcommand{\yes}{\ding{51}}
\newcommand{\no}{\ding{55}}
\definecolor{wipcol}{HTML}{E07B00}

\title{Fork Where the Model Changes Its Mind:\\ Belief-Shift Branching for Tree-Structured Reinforcement Learning}

\newif\ifpreprint
\preprinttrue
\author{Bin Lei$^{1,2}$ \quad Yu Li$^{2}$ \quad Prafulla Kumar Choubey$^{2}$ \quad
Jiaxin Zhang$^{2}$ \quad Becky Xiangyu Peng$^{2}$ \\
\textbf{Qinyuan Ye}$^{2}$ \quad \textbf{Kartik Narayan}$^{2}$ \quad \textbf{Caiwen Ding}$^{1}$ \quad \textbf{Silvio Savarese}$^{2}$ \quad
\textbf{Chien-Sheng Wu}$^{2}$ \\
$^{1}$University of Minnesota \quad $^{2}$Salesforce AI Research \\
{\fontsize{8.5}{9.5}\fontfamily{zi4}\selectfont{\{bin.lei,yu.li,pchoubey,jiaxin.zhang,becky.peng\}@salesforce.com}} \\
{\fontsize{8.5}{9.5}\fontfamily{zi4}\selectfont{\{qinyuan.ye,kartik.narayan,ssavarese,wu.jason\}@salesforce.com} \quad {\{lei00126,dingc\}@umn.edu}}}

\ifpreprint\iclrfinalcopy\fi
\begin{document}

\maketitle
\ifpreprint\lhead{Preprint}\fi

\begin{abstract}
\input{sections/0_abstract}
\end{abstract}

\input{sections/1_intro}
\input{sections/2_background}
\input{sections/3_method}
\input{sections/4_prerl}
\input{sections/5_experiments}
\input{sections/6_rlresults}
\input{sections/6b_ablations}
\input{sections/9_conclusion}
\newpage

\subsection*{AI use statement}
In this work, we used generative AI tools for writing assistance (editing prose
and \LaTeX{} formatting), for plotting and analysis scripts, and for
experiment-orchestration scripts (job scheduling and monitoring). The research
ideas, method design, and experimental conclusions are the authors' own; all
AI-assisted code was reviewed and tested by the authors, and all numbers
reported in the paper were produced by our training and evaluation pipeline and
verified against raw logs. Two uses go beyond assistance and are part of the
method or its evaluation: the surge/steady/drop continuations used to fit the
BSV direction (Appendix~\ref{app:bsvdata}) were drafted by Claude Opus 4.8 from
the probe models' own rollouts and adversarially checked by a second Claude
Opus 4.8 pass before acceptance, and the black-box
LLM-judge baseline of the pre-RL validation (Section~\ref{sec:prerl}) is
GPT-5.5. Neither enters any RL training run. We take responsibility for the
final content of this work, including text, claims, and artifacts produced with
the aid of generative AI.

\subsection*{Ethics statement}
This work studies reinforcement learning for mathematical and coding reasoning
on publicly available models and datasets. It involves no human subjects and no
personally identifiable data. Improved reasoning ability carries the usual
dual-use considerations of stronger language models; our method does not
introduce risks beyond those of standard RLVR training.

\subsection*{Reproducibility statement}
Section~\ref{sec:experiments} specifies models, datasets, tree configuration,
and optimization hyperparameters; Appendix~\ref{app:hyperparams} lists the
complete configuration, including the boundary-pool construction and probe
settings. All training uses publicly released base models and public datasets.
Source code will be released upon publication.

\bibliography{references}
\bibliographystyle{iclr2027_conference}

\appendix
\input{sections/10_appendix}

\end{document}

%% file: math_commands.tex
\usepackage{amsmath,amsfonts,bm}

\def\eqref#1{equation~\ref{#1}}

\def\1{\bm{1}}

\DeclareMathAlphabet{\mathsfit}{\encodingdefault}{\sfdefault}{m}{sl}
\SetMathAlphabet{\mathsfit}{bold}{\encodingdefault}{\sfdefault}{bx}{n}



%% file: sections/0_abstract.tex
Tree-structured rollouts give critic-free reinforcement learning with
verifiable rewards (RLVR) step-level credit: fork a
chain at an intermediate point, and sibling outcome differences estimate
step value. Each fork adds sampling cost, so realistic budgets typically allow only a few forks per chain. A fork placed where the outcome is already largely settled yields siblings that mostly agree and provide almost no credit signal; hence, for a given tree size, where forks are placed largely determines how much step-level RL can gain. Most existing mainstream methods place forks by structure, such as fixed lengths, midpoints, and delimiters, or by next-token entropy. We formalize fork placement as locating the
\emph{pivots} of the chain's value curve, where the expected outcome
turns. We propose \emph{belief-shift branching}: read the
model's answer belief at candidate boundaries and fork just before the
step where consecutive beliefs diverge most. Three
instantiations, none needing step-level supervision, span access levels: a black-box probe, a
logit-lens depth profile, and a learned activation direction, which is
fit offline and therefore used only in the validation before RL training. The signal only
\emph{places} forks, and the probe costs about $1\%$ of step compute on
mathematics and under $5\%$ on code when it runs inside the rollout engine. In that validation, against Monte-Carlo value curves, a belief-shift
signal ranks first in each of the eight model$\times$benchmark panels,
ahead of entropy, structural, and LLM-judge baselines. In RL across three model families and
two domains, belief-shift forking leads every
mathematics aggregate, on OLMo-3-7B by $+2.6$ aggregate and $+2.9$ on
AIME 2026 over the strongest baseline, and sweeps every OLMo code column,
by $+6.5$ on LiveCodeBench-medium.

%% file: sections/1_intro.tex
\section{Introduction}
\label{sec:intro}

\suppressfloats[t]
\begin{figure}[t]
\centering
\includegraphics[width=0.92\linewidth]{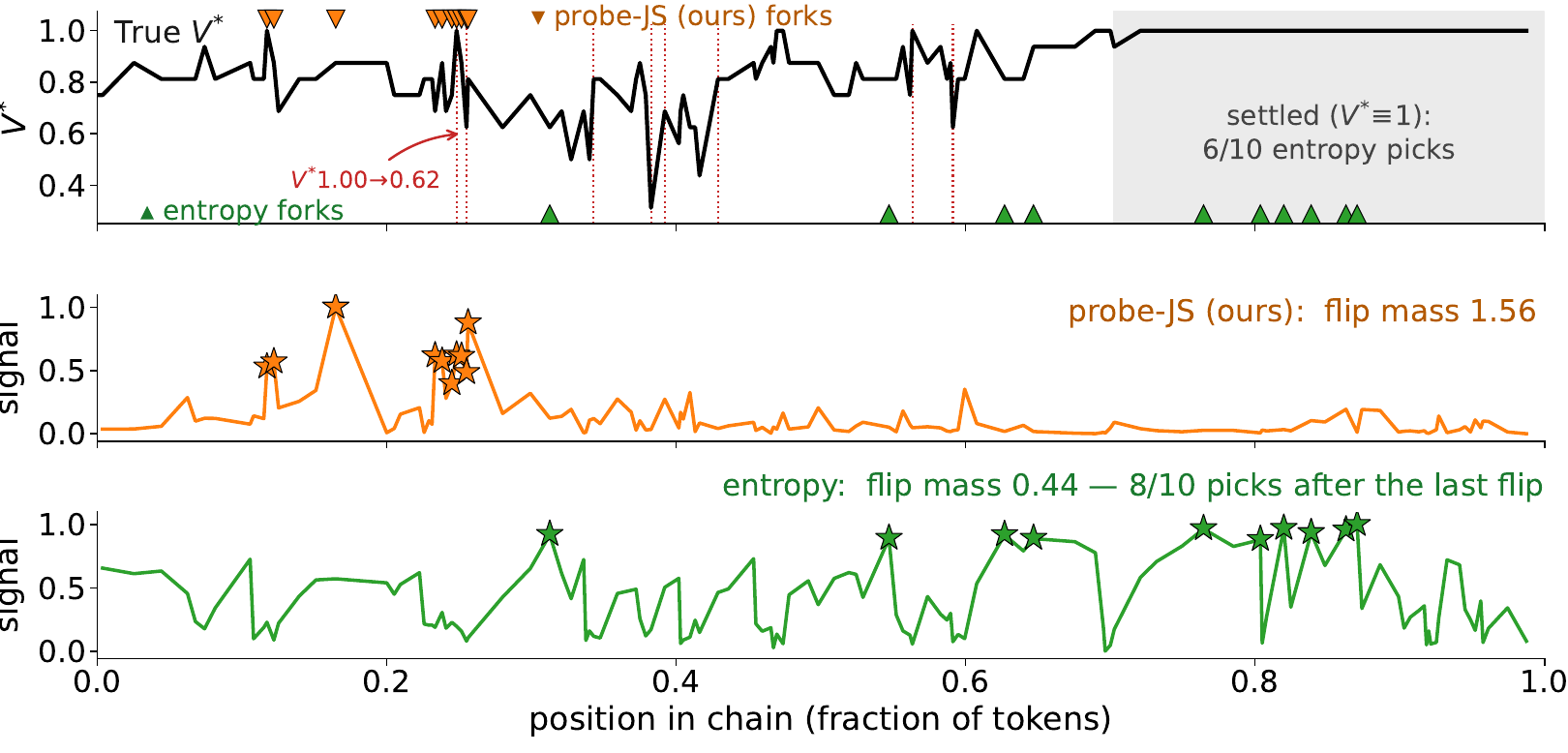}
\caption{An OLMo-3.1-32B-Think chain on AIME~2025; 128 newline
boundaries, ten forks per selector; $x$-axis: chain position (fraction).
\emph{Top}: ground-truth $V^\ast$ (16 MC completions per boundary); red
dotted: the eight value pivots with $|\Delta V^\ast|\!\ge\!0.25$; shaded:
settled region ($V^\ast\!\equiv\!1$); stars: fork positions. Flip mass
captured by the ten forks, $\sum_{\text{forks}}|\Delta V^\ast|$:
\probejs{} $1.56$ vs.\ entropy $0.44$, which spends six forks in the
settled region. Full case: Appendix~\ref{app:introcase}.}
\label{fig:selectors}
\end{figure}

Reinforcement learning with verifiable rewards (RLVR) trains reasoning
models with one scalar per sampled
solution~\citep{guo2025deepseek,shao2024deepseekmath}. Group-relative
methods copy that scalar onto every token, mis-crediting at both ends: a
flawed step inside a luckily-correct solution is reinforced in full, and a prompt whose
rollouts all fail yields zero
gradient~\citep{yu2026dapo,uesato2022solving}. Tree-structured rollouts
repair this without a critic: fork the trajectory at an intermediate
position, and the sibling continuations' verified outcome difference is a counterfactual estimate of the forked
step's advantage~\citep{hou2025treerl,yang2025treerpo,li2025treepo}. The
The catch is cost: each fork adds rollout tokens, so realistic budgets typically allow only a few forks per chain.

Where should those scarce forks go? In critic-free tree RLVR this is an open design choice: the tree already converts sibling outcomes
into step credit, so what remains is deciding which step the siblings
should re-roll. Current selectors make this choice without asking what
the step \emph{decides}: TreePO and TreeRPO fork on a fixed token
grid~\citep{li2025treepo,yang2025treerpo}, earlier step methods cut at
delimiters~\citep{lightman2024let}, TreeRL and FR3E fork at next-token uncertainty
peaks, and ARPO at post-tool entropy
rises~\citep{hou2025treerl,zheng2025first,dong2026agentic}. Uncertainty is
the natural candidate, but it tracks freedom of \emph{wording} rather
than of \emph{outcome}: entropy is inflated by interchangeable
paraphrases~\citep{kuhn2023semantic}, and it reads near zero for any
confidently written step, right or wrong, whereas an answer probe moves
when the step changes the model's own answer. Consistent with this, TreeRL's
surprisal-based forking (its ``entropy'') beats \emph{random} forking by
$2.1$ pass-rate points in its own sampling ablation, and TreePO's static
probability-guided branching allocation loses to uniform.

We propose belief-shift branching: track the model's belief over the final
answer along the chain and fork where that belief shifts most between
consecutive candidate boundaries. We instantiate it with three reads spanning
access levels---from output-only sampling to internal activations---none
needing step-level supervision. PROBE-JS is a black-box read in the spirit of
a pop quiz: at each boundary it interrupts the model with a $\sim$10-token
probe (``so what is your final answer?''), elicits the answer distribution,
and scores the boundary by the Jensen--Shannon divergence between consecutive
distributions, so that a sharp jump marks a pivot or a mistake; it needs only
sampling access, at the cost of a short extra generation per boundary.
LENS-SHIFT is a white-box read---an X-ray rather than a quiz---that asks the
same question without making the model speak: given the final answer, the
logit lens reads how strongly each layer already commits to it, yielding a
depth profile at every boundary, and a large change in this profile between
boundaries marks a belief shift, at no extra generation. Belief-Shift Vector
(BSV) is a radar calibrated in advance: by contrasting activations under high
confidence with those under doubt, it fits offline, before RL, a single
activation direction for ``confidence shift'', so scoring a boundary is just a
projection of its hidden state onto this axis. Since BSV measures belief shift
most directly, we use it to test whether, if the model's true belief shift
could be measured directly, it would track the ground-truth value curve more
accurately.

The probe costs about $1\%$ of a training step on mathematics and under
$5\%$ on code when served by the rollout engine (Section~\ref{sec:cost}), and the
signal only \emph{places} forks. Credit still comes from verified
sibling outcomes, so a miscalibrated belief wastes a fork but cannot
corrupt training. Figure~\ref{fig:selectors} contrasts \probejs{} with
entropy on a real chain against a Monte-Carlo ground-truth value curve:
given ten forks each, the probe's forks capture $3.6\times$ the value
movement of entropy's ($\sum|\Delta V^\ast|$: $1.56$ vs.\ $0.44$), and
entropy wastes six forks in the settled region where $V^\ast\!\equiv\!1$.

\emph{Before} RL, we score selectors against Monte-Carlo value
curves: a belief-shift signal ranks first in each of the eight
model$\times$benchmark panels, ahead of entropy, structural, and LLM-judge
baselines, and transfers zero-shot to an out-of-domain benchmark
(Section~\ref{sec:prerl}). \emph{During} RL, four
arms that differ only in the fork criterion span three model families
and two domains under matched budgets: a belief-shift arm leads every
mathematics aggregate ($+2.9$ on OLMo-3-7B AIME 2026 over the strongest
baseline) and sweeps every OLMo code column (Section~\ref{sec:results}). Ablations cover the
tree budget $(M,k)$, policy updater, probe read length $L$, and \lens{}
read-out, all on OLMo-3-7B mathematics, plus model scale on Qwen3-4B
versus Qwen3-8B (Section~\ref{sec:ablations}).

%% file: sections/2_background.tex
\section{Background and Related Work}
\label{sec:background}

\paragraph{Trajectory-level RLVR and its blind spots.}
\label{sec:trajrl}
Group-relative RLVR~\citep{shao2024deepseekmath,yu2026dapo} gives every
token of a response the same advantage $\hat A_i\propto R_i-\bar R$. Two
pathologies follow: final-answer rewards leave far more correct
solutions with flawed reasoning than process
feedback~\citep{uesato2022solving}, and all-tie prompts yield zero
gradient~\citep{yu2026dapo}. Process supervision improves on
outcome-only training in reasoning
settings~\citep{wang2024math,yuan2024free,setlur2025rewarding},
with up to \mbox{$6\times$} better RL sample efficiency from process
advantage verifiers~\citep{setlur2025rewarding}.

\paragraph{Three rollout structures, one common input.}
\label{sec:paradigms}
Absent any learned value estimator (whether an explicit critic or an
implicit one recovered from outcome-trained
log-ratios~\citep{yuan2024free,cui2025process}), step values must come from
extra sampling:
(a)~Monte-Carlo completions from step
prefixes~\citep{kazemnejad2024vineppo,wang2024math,guo2026segment};
(b)~MCTS-style trees, in practice used to label process supervision offline
or alongside learned value
models~\citep{guan2025rstar,zhang2024rest,luo2024improve,chen2024alphamath,feng2023alphazero};
(c)~on-policy branching trees whose node statistics directly yield RL
advantages without any value
model~\citep{hou2025treerl,yang2025treerpo,li2025treepo,dong2026agentic}, our
setting. All three consume the same input: boundaries where a chain may be
cut, of which realistic budgets afford only a few per chain.

\begin{table}[t]
\caption{\textbf{Existing fork-point selectors.} Few read the outcome at fork time: grids and delimiters follow surface
structure, surprisal and entropy track wording freedom, and the LM judge
reads correctness but needs an external generation call per step.}
\label{tab:selectors}
\begin{center}
\scriptsize
\setlength{\tabcolsep}{3.5pt}
\begin{tabular}{lllcl}
\toprule
Fork selector & Signal read & What it tracks & Outcome & Extra cost \\
\midrule
Fixed grid~\citep{li2025treepo,yang2025treerpo} & token count & none (content-blind) & \no & $0$ \\
Delimiters~\citep{lightman2024let,kazemnejad2024vineppo} & surface markers & format convention & \no & $0$ \\
Surprisal~\citep{hou2025treerl} & $-\log p$ of sampled token & wording surprise & \no & $0$ (reuses rollout) \\
Entropy~\citep{zheng2025first,dong2026agentic} & next-token distribution & wording freedom & \no & $0$ (reuses rollout) \\
LM judge~\citep{lai2024step} & external critique & judged correctness & (\yes) & ${\ge}1$ gen.\ call / step \\
\bottomrule
\end{tabular}
\end{center}
\end{table}

\paragraph{Existing fork selectors are mostly outcome-agnostic proxies.}
\label{sec:branchproblem}
Let $V^\ast(t)=\Pr[\text{correct}\mid x,y_{\le t}]$; sibling comparison at
$t$ estimates the local change of $V^\ast$, so the ideal selector targets
the (unobservable) pivots of $V^\ast$. Practice approximates them with
proxies that never read the outcome (Table~\ref{tab:selectors}). \emph{Delimiters}: PRM800K
fine-tunes the generator to emit newline steps~\citep{lightman2024let},
and surface tokens do not typically mark true decision
points~\citep{liu2025adaptivestep}. \emph{Fixed-length
grids}~\citep{li2025treepo,yang2025treerpo} fork mid-expression; TreePO's
static probability-guided branching allocation over its fixed segments
loses to uniform (their \S4.4). \emph{Next-token
uncertainty} (TreeRL's sampled-token surprisal, FR3E's entropy, ARPO's
post-tool entropy~\citep{hou2025treerl,zheng2025first,dong2026agentic})
measures wording, not outcome (Section~\ref{sec:intro}), and beats
\emph{random} forking by $2.1$ pass-rate points in TreeRL's own sampling
ablation at the same tree configuration, with about $8\%$ fewer generated
tokens. SPO~\citep{guo2026segment} already moves Monte-Carlo cutpoints off
a fixed grid, placing them at low-probability tokens on the premise that a
boundary is worth sampling only where the value may change; we share the
premise, but its read is again the probability of the sampled token,
whereas ours is the model's belief over the answer. \emph{LM judges} are
offline and costly~\citep{lai2024step},
and learned rewards are exploitable during RL~\citep{guo2025deepseek};
\emph{agent turns}~\citep{zhou2024archer,putta2024agent} cost $K^{T}$
rollouts if branched exhaustively over $T$ turns.

%% file: sections/3_method.tex
\section{Belief-Shift Branching}
\label{sec:method}

\paragraph{Template.}
A fork-point selector is a scoring rule over candidate boundaries: given a
chain $y$ with pool $\mathcal{B}(y)$ (Section~\ref{sec:boundaries}), it
assigns a score $s_t$ to each boundary $t\in\mathcal{B}(y)$ and forks at
the top scorer $\hat t=\arg\max_{t} s_t$. Our selectors
all take one form: read the model's current \emph{answer belief} at each boundary, and score the
segment between consecutive boundaries by how far it moved that belief,
$s_{t^-}=D(b_{t^-},b_t)$, where $t^-$ is the boundary immediately
preceding $t$ in $\mathcal{B}(y)$ and $D$ is an instantiation-specific
measure of belief change (a divergence or a profile-change norm; the
learned read of Section~\ref{sec:bsv} instead projects the single boundary
state $h_t$ onto a fitted belief-shift direction). Note the \emph{fork-before}
attribution: the belief change is caused by the segment $(t^-\!,t]$, so the
score lands on $t^-$ and siblings re-roll exactly the belief-moving step.
Three instantiations vary only in how $b_t$ is read: \probejs{} needs
sampling access and \lens{} reads hidden states, while BSV amortizes the read into a direction fit offline, which confines it to pre-RL validation—testing whether, if the model's true belief shift could be measured directly, it would track the ground-truth value curve more accurately (Section~\ref{sec:bsv}).

\subsection{Candidate step boundaries}
\label{sec:boundaries}

Mid-word forks yield siblings that differ for trivial lexical reasons, so
candidates are restricted to \emph{step boundaries}: positions after a
chosen delimiter, here the line break (\texttt{\textbackslash n}), the
natural step separator in both tasks we test (mathematics and code). At most $C{=}16$ evenly spaced
candidates are scored per chain; a chain whose span has no boundary is
cut at the span midpoint instead, so it still forks. Only a chain that
cannot fork at all (fewer than two tokens past the last cut, or no
response budget left for a continuation) is replaced by a fresh root chain
(Section~\ref{sec:prelim}), so every arm always produces its full tree.

\begin{figure}[H]
\centering
\includegraphics[width=0.98\linewidth]{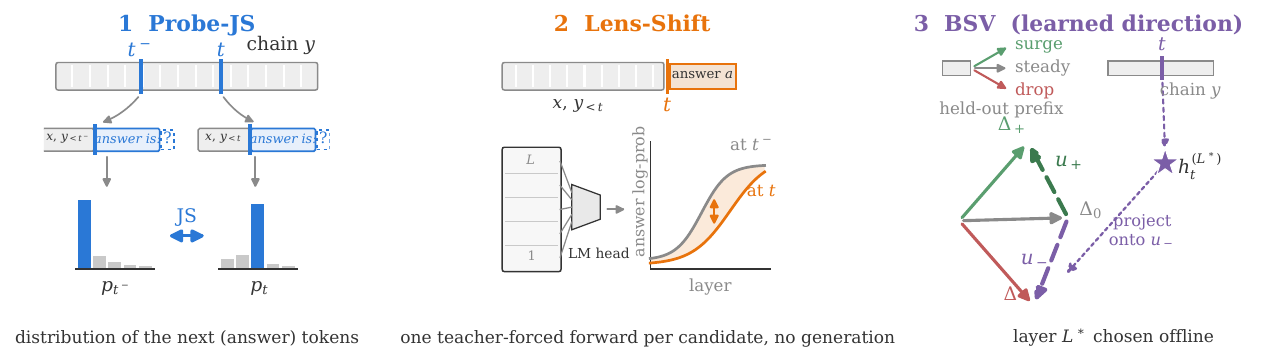}
\caption{The three belief reads: \probejs{} compares consecutive
elicited answer distributions (JS); \lens{} takes the $\ell_2$ change of
the teacher-forced answer's per-layer log-prob profile; BSV projects
$h_t$ onto an offline-fit surge/steady/drop direction (pre-RL only).}
\label{fig:methods}
\end{figure}

\subsection{\probejs{}: black-box belief probing}
\label{sec:probejs}

At boundary $t$ we append a short elicitation suffix $e$ that asks for
the final answer (``\texttt{So the final answer is}\,\dots'';
$E{\approx}8$--$9$ tokens) to the prefix
$y_{<t}$ and read the next-token distribution at the end of the suffix: the
belief $p_t$ is the top-$K$ ($K{=}20$) log-probabilities of
$\pi_\theta(\cdot \mid x, y_{<t}, e)$, i.e.\ the model's distribution over
the first answer token. Instantiating the template score
with the Jensen--Shannon divergence~\citep{lin1991divergence}:
\begin{equation}
s_{t^-} \;=\; \JS\!\left(p_{t^-}\,\|\,p_t\right)
\;=\;\tfrac12 D_{\mathrm{KL}}\!\left(p_{t^-}\|\,m\right)
   +\tfrac12 D_{\mathrm{KL}}\!\left(p_{t}\|\,m\right),
\qquad m=\tfrac{p_{t^-}+p_t}{2}.
\label{eq:probejs}
\end{equation}
Unlike KL, the mixture $m$ makes the score symmetric,
$\JS(p\|q)=\JS(q\|p)$, and bounded, $0\le\JS\le\log 2$: segments are
ranked by movement alone, on one scale, with no reference endpoint,
whereas $D_{\mathrm{KL}}(p_{t^-}\|\,p_t)\to\infty$ as $p_t(v)\to0$ for
any $v$ with $p_{t^-}(v)>0$, so under top-$K$ truncation a single tail token that
drops out of one support can own the argmax; the two-sided form also yields the
bound of Eq.~\ref{eq:certificate}. The truncation is an engineering constraint, not a modeling
choice: vLLM-class engines and OpenAI-style APIs return at most the
top-$K$ next-token log-probabilities, with $20$ the default
ceiling, hence $K{=}20$. With $S_t$ the returned support at
$t$ and $S=S_{t^-}\!\cup S_t$, each side is floored and renormalized on
$S$,
\begin{equation}
\tilde p_t(v)\;\propto\;
\begin{cases}
p_t(v), & v\in S_t,\\[1pt]
\tfrac12\,\min_{u\in S_t} p_t(u), & v\in S\setminus S_t,
\end{cases}
\label{eq:floor}
\end{equation}
and Eq.~\ref{eq:probejs} is evaluated on $(\tilde p_{t^-},\tilde p_t)$.
Each probe rides the rollout's prefix cache, forwarding only the suffix
plus the $L$ decoded tokens ($L{=}1$ for mathematics, $16$ for code;
ablated in Section~\ref{sec:ablations}). For code the suffix is
\texttt{So the final solution is:} followed by an opening \texttt{python}
fence, so the ``answer'' is the opening of the solution program. For
$L{>}1$ the engine samples $L$ tokens after the suffix at temperature $1.0$
and returns the top-$K$ distribution at each position; Eq.~\ref{eq:probejs}
is applied position by position between consecutive boundaries and
averaged over the $L$ positions, so $L{=}1$ is the special case above.
Position one is the exact $L{=}1$ read; positions $2$ to $L$ are
conditioned on that boundary's own sampled read-out, so the score adds a
sampled component on top of the $L{=}1$ signal. The same read defines the
$L$ ablation of Section~\ref{sec:ablations}.

\subsection{\lens{}: white-box belief reading}
\label{sec:lens}

With activation access, beliefs are read from inside the forward pass. The
logit lens~\citep{nostalgebraist2020interpreting} defines per-layer
predictions
$P^{(\ell)}(\cdot)=\mathrm{softmax}\!\big(W_U\,\mathrm{LN}(h^{(\ell)})\big)$,
where $h^{(\ell)}$ is the residual stream after layer $\ell$ and
$\mathrm{LN}$, $W_U$ are the model's own final norm and unembedding.
After the prefix $y_{<t}$ and the elicitation suffix of
Section~\ref{sec:probejs}, we teacher-force the chain's committed answer
$a=(a_1,\dots,a_n)$ (its final answer, $n$ tokens) and record how
strongly each intermediate layer already encodes that answer (no
sampling is involved), giving a \emph{depth profile} and
its change score
\begin{equation}
b_t^{(\ell)} \;=\; \sum_{i=1}^{n}\log
P^{(\ell)}\!\left(a_i \,\middle|\, x,\,y_{<t},\,e,\, a_{<i}\right),
\qquad
s^{\mathrm{shape}}_{t^-}=\bigl\|\,b_t^{(\cdot)}-b_{t^-}^{(\cdot)}\bigr\|_2 ,
\label{eq:lensscore}
\end{equation}
with $x$ the prompt and $b_t^{(\cdot)}$ the vector collecting
$b_t^{(\ell)}$ over all probed layers. The profile traces the answer's
formation across depth, so its change localizes \emph{where} a step
moved the belief; $s^{\mathrm{shape}}$ fires when that
path reorganizes at any depth. Two scalar read-outs,
\begin{equation}
s^{\mathrm{height}}_{t^-}=\bigl|\overline{b_t}-\overline{b_{t^-}}\bigr|,
\quad
s^{\mathrm{depth}}_{t^-}=\bigl|\ell^\ast_t-\ell^\ast_{t^-}\bigr|,
\quad
\ell^\ast_t=\min\Bigl\{\ell:\,
\bigl|b^{(\ell)}_t-b^{(\ell_{\max})}_t\bigr|\le
\varepsilon\cdot\operatorname{range} b^{(\cdot)}_t\Bigr\},
\label{eq:lensvariants}
\end{equation}
($\overline{b}$: profile mean over the deepest half of the layers;
$\ell_{\max}$: last layer; $\varepsilon{=}0.1$) read the settled
belief's strength and its lock-in depth. All RL arms use
$s^{\mathrm{shape}}$ ($s^{\mathrm{height}}$/$s^{\mathrm{depth}}$: the
read-out ablation of Section~\ref{sec:ablations}); the cost is one
teacher-forced forward per candidate, no generation.

\subsection{Belief-shift vector: a learned read for pre-RL validation}
\label{sec:bsv}

BSV is not an RL fork selector but a validation instrument: it tests whether forking at the model's true belief shifts, read as directly as activation access allows, reconstructs the ground-truth value curve more accurately than the proxies of Table~\ref{tab:selectors}. To make that read as direct as possible, we amortize the per-boundary belief read into one dot product (Figure~\ref{fig:methods}), fit once, offline, on a \emph{frozen} policy. A
held-out chain is cut at a boundary $t$ and continued in three
LLM-drafted continuation regimes (Appendix~\ref{app:bsvdata}), indexed $g\in\{+,0,-\}$: \emph{surge} (the
continuation commits to the answer), \emph{steady} (reasoning proceeds
with no change of belief), and \emph{drop} (it retracts the answer or expresses doubt). Per
layer $L$ and regime $g$, the residual stream is averaged over a
$W_a{=}16$-token window after the cut and a $W_b{=}16$-token window
before it and differenced, then the steady-regime delta is subtracted:
\begin{equation}
\Delta^{(L)}_{g}=\mathop{\mathbb{E}}_{(t,c):\,g(c)=g}\Bigl[\bar h^{(L)}_{[t,\,t+W_a)}(c)-\bar h^{(L)}_{[t-W_b,\,t)}\Bigr],
\quad
u^{(L)}_{\pm}=\Delta^{(L)}_{\pm}-\Delta^{(L)}_{0}.
\label{eq:bsvfit}
\end{equation}
$\Delta_0$ removes the progress shared by all continuations, leaving pure
belief-change axes $u_{\pm}$; their antagonism picks the layer label-free,
and scoring is one projection of the last prefix token:
\begin{equation}
L^\ast=\arg\min_{L}\cos\bigl(u^{(L)}_{+},u^{(L)}_{-}\bigr),
\qquad
s_{t}=\bigl\langle u^{(L^\ast)}_{-},\,h^{(L^\ast)}_{t}\bigr\rangle .
\label{eq:bsvscore}
\end{equation}
The fit is also what keeps BSV out of the RL arms: Eq.~\ref{eq:bsvfit} averages
activation deltas of the policy that generated them, so once
training moves $\theta$ the direction is off-policy, scored against
activations the fit never saw, and refreshing it means re-running the
fork/group/average pipeline at training cadence, a cost far beyond the probes'. We therefore use BSV only where the
policy is frozen, as pre-RL evidence that forking at belief shifts
reconstructs the true value curve more accurately than every baseline (Section~\ref{sec:prerl}); the two fit-free reads carry the RL arms
(Appendix~\ref{app:extra}).

\subsection{Integration into tree-structured RLVR}
\label{sec:prelim}

We plug the selectors into a one-fork-per-chain instantiation of TreeRL's
on-policy tree scheme~\citep{hou2025treerl}: per prompt,
$M{=}4$ chains are sampled and each forks at its top-scoring boundary into
$k{=}2$ sibling continuations, giving 12 scored rollouts per prompt (a chain
that cannot fork at all, under two tokens or out of response budget, is
replaced by a fresh root chain, so each prompt always emits 12 rollouts).
Once a chain has finished generating, the selector scores its boundaries
(for \probejs{}, with independent one-token probe requests on the finished
chain's prefix), and the siblings are ordinary engine requests on that
prefix: with automatic prefix caching the prefix's KV blocks are reused
and only the continuation is generated, which is what the rollout term of
Eq.~\ref{eq:cost} charges. On the training side the 12 rows are processed
as separate sequences, prefix included; the $T_{\text{train}}$ of
Appendix~\ref{app:cost} is measured that way. A verifier scores each
leaf, lightly shaped by a DAPO-style overlong penalty~\citep{yu2026dapo}.
Advantages follow TreeRL's global-plus-local form: for a node with
subtree-mean return $V$, sibling-group mean $\mu$, tree mean $\bar V$, and
$|\mathcal{L}|$ descendant leaves,
\begin{equation}
A \;=\; \bigl[(V-\bar V)+(V-\mu)\bigr]\big/\sqrt{|\mathcal{L}|},
\label{eq:treeadv}
\end{equation}
broadcast to the segment's tokens. The optimizer is the standard RLVR
composite (group-relative policy gradient, asymmetric clipping
$(0.2,0.28)$, token-mean aggregation, dynamic
sampling~\citep{shao2024deepseekmath,yu2026dapo,liu2025understanding}),
identical across arms.
Baselines: \midpoint{} cuts the remaining span in half; \entropybl{}
(adapted from TreeRL, which ranks single tokens) forks where the segment-mean sampled-token surprisal
$-\log\pi(y_t)$ peaks over the same candidate pool.

\subsection{Cost and a probe-value bound}
\label{sec:cost}

\begin{table}[t]
\caption{\textbf{Probe overhead}: FLOP upper bounds at $C{=}16$ candidates and
$10$ / $25$ tokens per engine probe ($L{=}1$ / $16$; cells read mathematics /
code), with measured mean chain lengths (Nemotron's mathematics $\bar T$
inferred from ${\sim}660$k rollout tokens per step).}
\label{tab:cost}
\begin{center}
\small
\begin{tabular}{lcc}
\toprule
Model & Mean chain length $\bar T$ & Probe FLOPs / full training step \\
\midrule
OLMo-3-7B    & $1{,}327$ / $1{,}179$ & $1.3\%$ / $3.2\%$ \\
Qwen3-4B     & $4{,}164$ / $760$ & $0.45\%$ / $4.3\%$ \\
Nemotron-9B  & $\sim2{,}600$ / $6{,}557$ & $0.70\%$ / $0.67\%$ \\
\bottomrule
\end{tabular}
\end{center}
\vspace{-8pt}
\end{table}

\paragraph{Cost.}
Let $N$ be the parameter count, $P$ the prompts per step, $\bar T$ the
mean chain length in tokens, and $M$, $k$ the root chains and siblings
per cut of Section~\ref{sec:prelim}. With inference at $2N$ FLOPs/token,
rollout costs $F_{\mathrm{roll}}\!\approx\!2N\,PM\,[1{+}k(1{-}c)]\,\bar T$
(each chain plus its $k$ siblings, which continue from a cut at fraction
$c$ of the chain, $c\approx0.5$). Probing $C$
candidates with suffix length $E$ and read length $L$ on the engine's
prefix cache costs $F_{\probejs}=2N\,PMC(E{+}L)$, giving
\begin{equation}
\frac{F_{\probejs}}{F_{\mathrm{roll}}}
= \frac{C(E{+}L)}{[1{+}k(1{-}c)]\,\bar T}
\;\xrightarrow[\;\bar T\to\infty\;]{}\;0 .
\label{eq:cost}
\end{equation}
The probe term carries only the fixed probe length $E{+}L\ll\bar T$, so
the probe is a vanishing fraction of rollout, smaller still of the full
step, and \emph{shrinks} as chains lengthen. Equation~\ref{eq:cost} is a
per-chain upper bound ($C$ is a cap on the candidates actually scored, and
every generated prompt is probed); Table~\ref{tab:cost} reports the probe's share of a full
training step under the accounting of Appendix~\ref{app:cost}. We report FLOPs rather than wall-clock because the
arms are not wall-clock comparable: the probe path and the serving engine
differ across models (Appendix~\ref{app:cost}). \lens{} reads $n$ answer tokens
instead of $L$ and with prefix reuse costs the same order; ours
re-forwards each prefix on a co-located copy (the engine hides
intermediate layers), $C\rho\bar T/2$ token-forwards per chain
(Appendix~\ref{app:cost}).

\paragraph{Theoretical analysis: what belief shift can see.}
Let $V^\ast(t)$ be the true value of the prefix $y_{\le t}$ (the success
probability of the policy's own continuations, the Monte-Carlo quantity
of Section~\ref{sec:prerl}) and $V_e(t)=p_t(a^\ast)$ the \emph{probe
value}: the mass the elicited belief of Section~\ref{sec:probejs} puts on
the first token $a^\ast$ of the correct answer when the suffix $e$ forces an
answer at $t$ (for $L{>}1$, averaged over the $L$ read positions).
The two are not equal ($V_e$ makes the model answer \emph{now}; $V^\ast$
lets it keep reasoning), so $V_e$ is an approximation of $V^\ast$ whose
fidelity we measure rather than assume (reconstruction of Monte-Carlo
$V^\ast$, Section~\ref{sec:prerl}). For the probe value the guarantee is
exact. With
$\mathrm{TV}(p,q)=\tfrac12\sum_a|p(a)-q(a)|=\max_{A}|p(A)-q(A)|$ and
Pinsker's inequality on both halves of Eq.~\ref{eq:probejs},
$\JS\ge\tfrac12\mathrm{TV}^2$, hence
\begin{equation}
\bigl|V_e(t)-V_e(t^-)\bigr|
\;\le\; \mathrm{TV}\!\left(p_{t^-},p_t\right)
\;\le\; \sqrt{2\,\JS\!\left(p_{t^-}\|\,p_t\right)}
\;=\;\sqrt{2\,s_{t^-}}
\label{eq:certificate}
\end{equation}
(proof in Appendix~\ref{app:theory}), i.e.\ the \probejs{} score
lower-bounds the movement of the probe value: a near-zero score cannot
hide a moved $V_e$. How closely $V_e$ tracks $V^\ast$ is measured, not
assumed, by the reconstruction results of Section~\ref{sec:prerl}.

%% file: sections/4_prerl.tex
\section{Pre-RL validation: do the signals find the true value pivots?}
\label{sec:prerl}

End-to-end RL is a confounded instrument for judging a branching signal: optimizer noise, seed variance, and the interaction between reward shaping and advantage estimation can all mask the signal's own effect. To isolate that effect, we first measure every selector directly against the ground-truth value curve, before any training.

\paragraph{Setup.}
The probe suites are AIME 2025--2026 and
GPQA-Diamond~\citep{rein2023gpqa}. For each of four probe
models (Gemma-4-E4B, OLMo-3-7B-Think, Gemma-4-31B, and
OLMo-3.1-32B-Think~\citep{team2026gemma,olmo2025olmo}; Hugging Face
ids in Table~\ref{tab:models}) we sample one
chain per problem at temperature $0.7$, cut it at every newline, and
evenly subsample the cuts to at most $m{=}128$ candidate boundaries
$t_1,\dots,t_m$ (all inference settings: Table~\ref{tab:prerlconfig},
Appendix~\ref{app:hyperparams}).

\paragraph{Ground truth and metric.}
The value curve $V^\ast(t_j)=\Pr[\text{correct}\mid y_{\le t_j}]$ is
estimated at every boundary by $N_{\mathrm{MC}}{=}16$ independent
Monte-Carlo completions of the prefix, scored by the verifier. Each
selector then ranks the $m$ boundaries by its own score; we keep its top
$B{=}10$, reconstruct $\hat V$ by linear interpolation through the kept
points, and charge the selector the reconstruction error
$E=\sum_j\lvert V^\ast(t_j)-\hat V(t_j)\rvert$ (lower is better), which is low
exactly when the chosen boundaries bracket the true pivots. Flat-$V^\ast$
problems are excluded; a dynamic program over $V^\ast$ gives the per-panel oracle floor.

\paragraph{Selectors compared.}
Ours: BSV (vectors fit on held-out HMMT~\citep{balunovic2026matharena};
layer by the label-free $\cos$ criterion of Section~\ref{sec:bsv}), \probejs{},
and \lens{} (shape/height). BSV's direction would need re-fitting at
every policy update, so it is a pre-RL instrument only; the two fit-free
reads carry the RL arms. Baselines: a \emph{black-box LLM judge}
(GPT-5.5 rates step importance from the chain alone), the only
baseline with content access, plus entropy, newline, uniform, and random
placement.

\begin{figure}[t]
\centering
\includegraphics[width=0.98\linewidth]{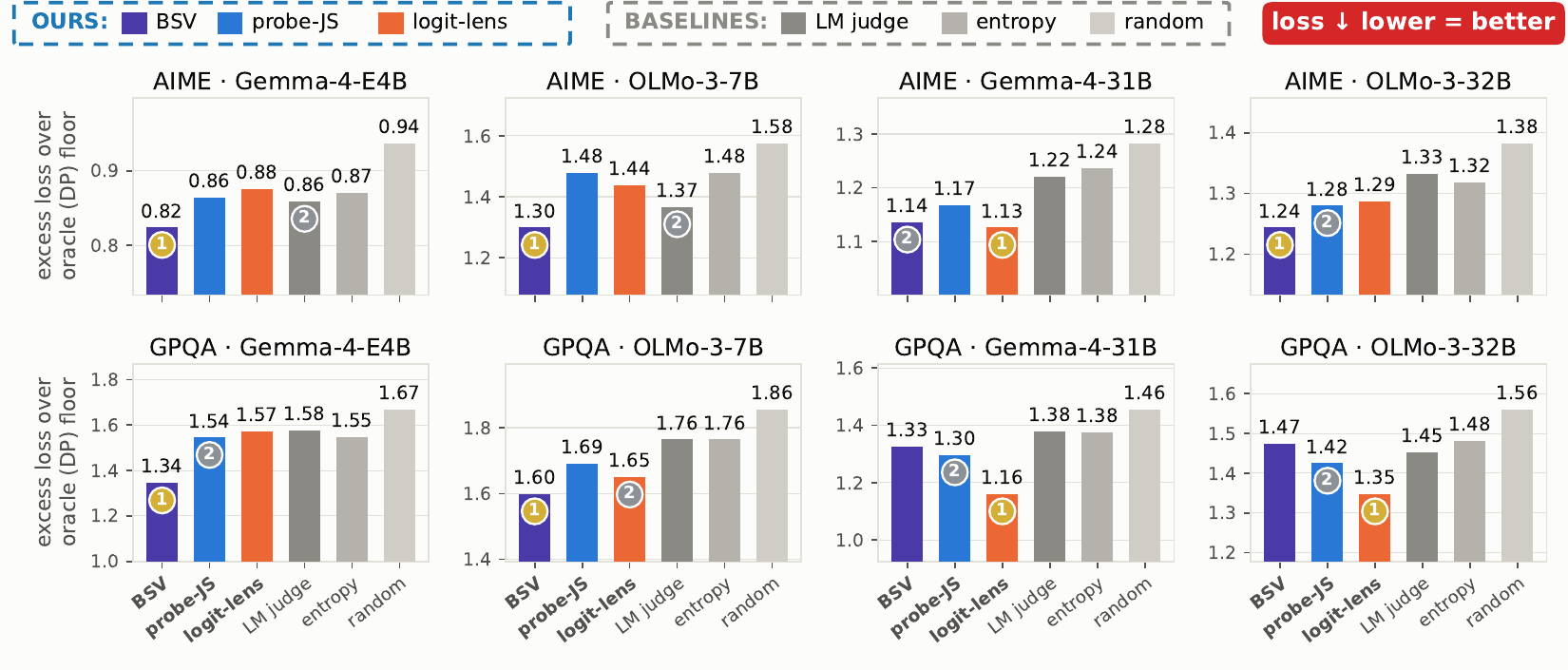}
\caption{\textbf{Pre-RL signal quality} (lower is better): reconstruction
error \emph{above the DP oracle floor} on AIME (top) and the held-out
GPQA-Diamond transfer (bottom), four probe models. A belief-shift selector
(colored) ranks first in each panel; BSV bars use the $\cos$-selected
layer (GPQA: fit out-of-domain on math); exact values in
Table~\ref{tab:prerl}.}
\label{fig:prerl}
\end{figure}

\paragraph{Results.}
Figure~\ref{fig:prerl} shows excess error over the oracle floor (exact
values: Table~\ref{tab:prerl}).
\emph{(i)}~A belief-shift signal takes the top rank in all eight
model$\times$benchmark panels. \emph{(ii)}~Entropy hovers at the level of blind newline placement.
The LLM judge, which reads the chain, matches the fit-free reads on AIME
and trails every belief-shift read on GPQA ($6.042$ vs.\ $5.938$--$5.988$),
at the cost of a generation call per step. \emph{(iii)}~The signal transfers: on
held-out GPQA-Diamond, BSV fit on \emph{out-of-domain math} beats every
baseline on average ($5.935$ vs.\ $6.041$) and in three of four panels,
and edges the in-domain fit ($5.999$): the belief direction is not
benchmark-specific.

BSV with the $\cos$-selected layer, fit only to \emph{locate} belief shifts
(outcome labels enter its fitting set only to balance correct and wrong
endings within every regime, never as a target; Appendix~\ref{app:bsvdata}),
beats every baseline in seven of the eight panels and on both averages, so localizing the shift suffices for a faithful $\hat V$. Its
remaining choice, which layer to project, is also label-free: the
antagonism criterion
$L^\ast=\arg\min_L \cos(u^{(L)}_+,u^{(L)}_-)$ correlates with
reconstruction quality on every model (Figure~\ref{fig:cosloss},
appendix), so directions, layer, and score all come from the model's own
rollouts.

%% file: sections/5_experiments.tex
\section{RL Experimental Setup}
\label{sec:experiments}

\paragraph{Models.}
Three open substrates spanning architectures:
\textbf{Qwen3-4B-Base}~\citep{yang2025qwen3}, \textbf{OLMo-3-7B}
(SFT checkpoint)~\citep{olmo2025olmo}, and
\textbf{Nemotron-Nano-9B-v2-Base} (hybrid Mamba--Transformer)~\citep{basant2025nvidia}.

\paragraph{Domains and data.}
\emph{Mathematics}: training on DAPO-Math-17k~\citep{yu2026dapo};
validation on OlympiadBench (the 674-problem English text-only open-ended
maths split, mean@1)~\citep{he2024olympiadbench}, AIME 2026 (30 problems,
avg@16)~\citep{maa2025aime}, and Omni-MATH-500 (a 500-problem subset we sample from Omni-MATH,
mean@1)~\citep{gao2025omni}, plus their unweighted mean (\emph{Agg.}).
\emph{Competitive programming}: training on
DeepCoder-24K~\citep{luo2025deepcoder} (21.6K problems after removing
overlap with the LiveCodeBench-v6 window); validation on the 175 problems
added in LiveCodeBench-v6 (43/52/80 easy/medium/hard,
avg@8)~\citep{jain2025livecodebench} with a binary hidden-test pass reward.

\paragraph{Protocol.}
Four arms per model and domain, \midpoint{}, \entropybl{}, \probejs{},
and \lens{} (shape), share the tree construction of Section~\ref{sec:prelim}, the data
order, reward, and optimizer, and differ only in the fork criterion.
All arms are critic-free (GRPO-style); a learned value model is a
separate design axis and is not compared here. Nor do we re-test the tree
itself: TreeRL~\citep{hou2025treerl} established that its tree scheme beats
chain RL under matched rollout budgets, and every arm here shares that
scheme, so the comparison isolates a single variable, where the forks are
placed.
Validation (accuracy on the three suites) runs every 20 steps; each arm is reported at its
\emph{single best-aggregate checkpoint} (step in gray;
Appendix~\ref{app:hyperparams} reports every aggregate under two other
selection rules). Qwen arms are budget-matched to exactly 680 steps; OLMo and Nemotron arms
are capped at a common per-group budget, within which each arm may stop
earlier (per-arm step counts in Appendix~\ref{app:hyperparams}). Training uses verl~\citep{sheng2025hybridflow} with vLLM
rollouts~\citep{kwon2023efficient} on a single $8\times$H200 node (temperature $1.0$,
$\mathrm{lr}=10^{-6}$, $16\times12=192$ rollouts per step after dynamic
filtering). Complete hyperparameters in Appendix~\ref{app:hyperparams}.

%% file: sections/6_rlresults.tex
\section{RL Results}
\label{sec:results}

\begin{figure}[t]
\centering
\includegraphics[width=0.9\linewidth]{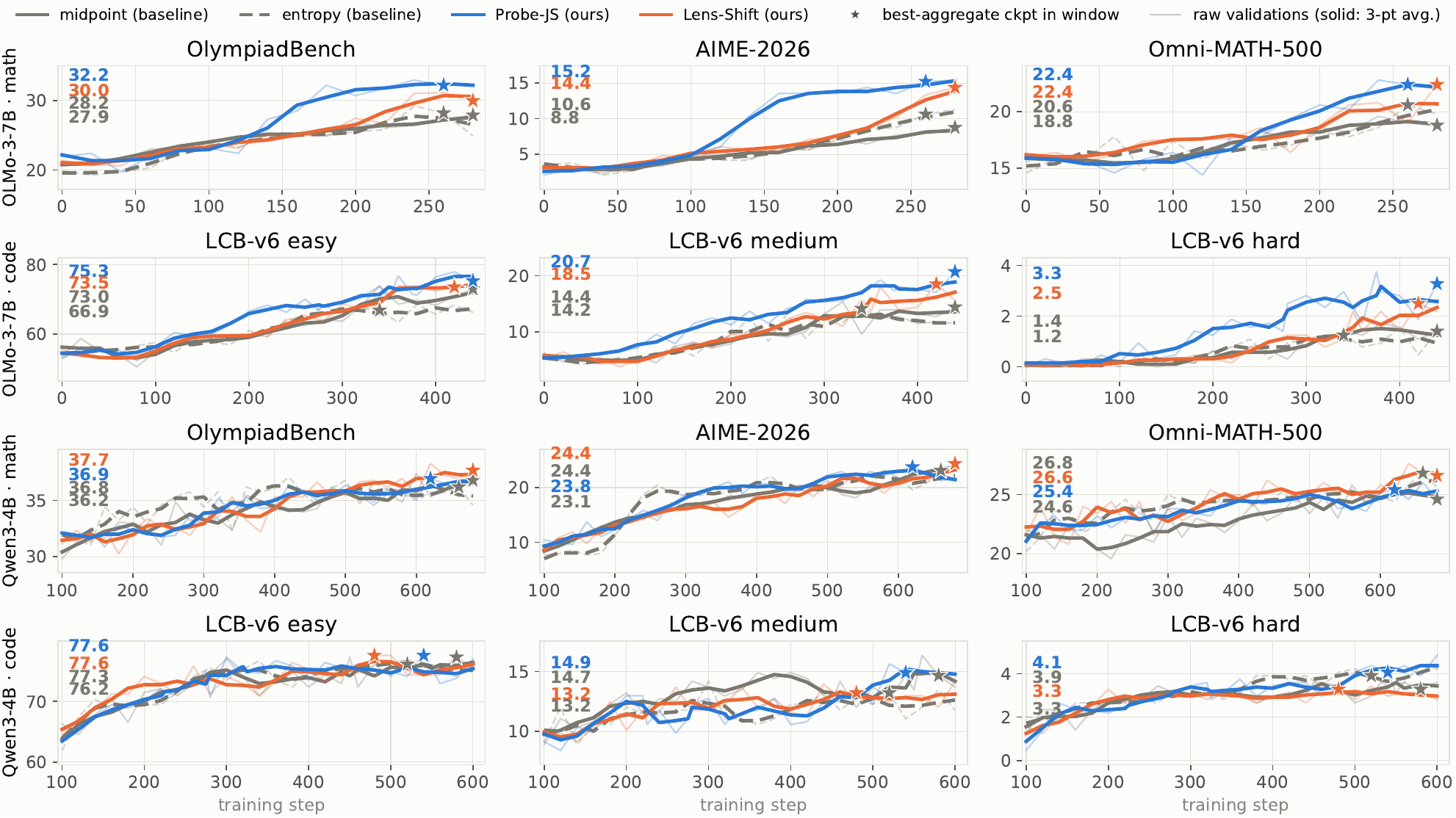}
\caption{\textbf{Per-benchmark validation accuracy} (\%), OLMo-3-7B
(rows 1--2) and Qwen3-4B (rows 3--4), mathematics and code. Solid:
3-point moving average; faint: raw validations every $20$ steps; $\star$
= best-aggregate checkpoint \emph{within the shown window} (value at the
left); Table~\ref{tab:math} follows the full-budget protocol of
Appendix~\ref{app:hyperparams}, so its checkpoint can lie beyond the
window. Windows: OLMo first $280$ / $440$ steps, Qwen steps
$100$--$680$ / $100$--$600$ (math / code; warm-up omitted).}
\label{fig:curves12}
\end{figure}

\begin{table}[t]
\caption{\textbf{RL results}: accuracy (\%) at each arm's \emph{single
best-aggregate checkpoint} (step in gray), so every row is read at a
single validation step. Mathematics: OlympiadBench, AIME 2026,
Omni-MATH-500 and their mean. Code (DeepCoder-24K $\to$
LiveCodeBench-v6): avg@8 on the easy / medium / hard splits and their
mean. Qwen arms are budget-matched to 680 steps; Nemotron code arms share a
$240$-step cap. \mdl{gold}{1}\,/\,\mdl{silver}{2} = best /
second best per column within each model block (ties share a medal).
Shaded rows: our belief-shift arms.}
\label{tab:math}\label{tab:code}
\begin{center}
\footnotesize
\setlength{\tabcolsep}{3.5pt}
\renewcommand{\arraystretch}{0.92}
\begin{tabular}{l cccc cccc}
\toprule
 & \multicolumn{4}{c}{\textbf{Mathematics}} & \multicolumn{4}{c}{\textbf{Code} (LiveCodeBench-v6)} \\
\cmidrule(lr){2-5}\cmidrule(lr){6-9}
Fork criterion & Olymp. & AIME & Omni & Agg. & easy & medium & hard & Agg. \\
\midrule
\multicolumn{9}{l}{\textit{OLMo-3-7B}} \\
\midpoint{}  & $28.2$ & $11.2$ & $19.0$ & $19.5$ \at{440} & \secondbest{75.0} & $16.1$ & $2.0$ & $31.0$ \at{480} \\
\entropybl{} & $28.6$ & $12.3$ & \secondbest{21.2} & $20.7$ \at{320} & $73.8$ & $12.0$ & $1.2$ & $29.0$ \at{460} \\
\cellcolor{ablpj}\probejs{}   & \cellcolor{ablpj}\best{32.2} & \cellcolor{ablpj}\best{15.2} & \cellcolor{ablpj}\best{22.4} & \cellcolor{ablpj}\best{23.3} \at{260} & \cellcolor{ablpj}\best{79.4} & \cellcolor{ablpj}\best{22.6} & \cellcolor{ablpj}\best{2.8} & \cellcolor{ablpj}\best{34.9} \at{520} \\
\cellcolor{abllens}\lens{}      & \cellcolor{abllens}\secondbest{30.0} & \cellcolor{abllens}\secondbest{14.4} & \cellcolor{abllens}\best{22.4} & \cellcolor{abllens}\secondbest{22.2} \at{280} & \cellcolor{abllens}$73.5$ & \cellcolor{abllens}\secondbest{18.5} & \cellcolor{abllens}\secondbest{2.5} & \cellcolor{abllens}\secondbest{31.5} \at{420} \\
\midrule
\multicolumn{9}{l}{\textit{Nemotron-9B}} \\
\midpoint{}  & $43.3$ & $47.1$ & $32.2$ & \secondbest{40.9} \at{340} & \best{100.0} & \best{63.5} & \best{30.2} & \best{64.5} \at{160} \\
\entropybl{} & $43.8$ & $42.9$ & $33.2$ & $40.0$ \at{200} & $98.5$ & $62.7$ & $27.5$ & $62.9$ \at{200} \\
\cellcolor{ablpj}\probejs{}   & \cellcolor{ablpj}\secondbest{44.3} & \cellcolor{ablpj}\best{55.6} & \cellcolor{ablpj}\secondbest{35.4} & \cellcolor{ablpj}\best{45.1} \at{320} & \cellcolor{ablpj}\secondbest{99.4} & \cellcolor{ablpj}\secondbest{63.2} & \cellcolor{ablpj}\secondbest{29.4} & \cellcolor{ablpj}\secondbest{64.0} \at{140} \\
\cellcolor{abllens}\lens{}      & \cellcolor{abllens}\best{44.8} & \cellcolor{abllens}\secondbest{53.8} & \cellcolor{abllens}\best{36.8} & \cellcolor{abllens}\best{45.1} \at{320} & \cellcolor{abllens}$98.0$ & \cellcolor{abllens}$63.0$ & \cellcolor{abllens}$27.5$ & \cellcolor{abllens}$62.8$ \at{200} \\
\midrule
\multicolumn{9}{l}{\textit{Qwen3-4B}} \\
\midpoint{}  & $36.8$ & \best{24.4} & $24.6$ & $28.6$ \at{680} & \secondbest{77.3} & \secondbest{14.7} & $3.3$ & \secondbest{31.8} \at{580} \\
\entropybl{} & $36.2$ & $23.1$ & \best{26.8} & \secondbest{28.7} \at{660} & \best{77.6} & $13.7$ & \secondbest{3.8} & $31.7$ \at{640} \\
\cellcolor{ablpj}\probejs{}   & \cellcolor{ablpj}\secondbest{36.9} & \cellcolor{ablpj}\secondbest{23.8} & \cellcolor{ablpj}$25.4$ & \cellcolor{ablpj}\secondbest{28.7} \at{620} & \cellcolor{ablpj}\best{77.6} & \cellcolor{ablpj}\best{14.9} & \cellcolor{ablpj}\best{4.1} & \cellcolor{ablpj}\best{32.2} \at{540} \\
\cellcolor{abllens}\lens{}      & \cellcolor{abllens}\best{37.7} & \cellcolor{abllens}\best{24.4} & \cellcolor{abllens}\secondbest{26.6} & \cellcolor{abllens}\best{29.6} \at{680} & \cellcolor{abllens}\best{77.6} & \cellcolor{abllens}$13.2$ & \cellcolor{abllens}$3.3$ & \cellcolor{abllens}$31.4$ \at{480} \\
\bottomrule
\end{tabular}
\end{center}
\end{table}

\subsection{Mathematics}
\label{sec:mathresults}

Figure~\ref{fig:curves12} shows the trajectories behind the OLMo and
Qwen rows of Table~\ref{tab:math}. A belief-shift arm takes the aggregate
lead on every model: \probejs{} on OLMo-3-7B ($23.3$ vs.\ $20.7$ for the
strongest baseline) and Nemotron-9B ($45.1$ vs.\ $40.9$), \lens{} on
Qwen3-4B ($29.6$ vs.\ $28.7$). On OLMo-3-7B, \probejs{} lifts
AIME 2026 from $12.3$ (strongest baseline) to $15.2$; entropy branching \emph{hurts} Nemotron ($40.0$ aggregate, below
midpoint), in line with the pre-RL finding that entropy does not track
value (Section~\ref{sec:prerl}). A belief-shift arm tops 11
of the 12 mathematics columns.

\subsection{Code}
\label{sec:coderesults}

On competitive programming (Table~\ref{tab:math}, right), \probejs{} wins every
OLMo-3-7B column ($34.9$ vs.\ $31.0$ for the best baseline, $+6.5$ on
LCB-medium; Figure~\ref{fig:curves12}, row 2). Qwen3-4B mirrors mathematics: \probejs{} posts the best aggregate
($32.2$ vs.\ $31.8$) and the best or joint-best score per split. Under the common $240$-step cap on Nemotron,
\midpoint{} leads and \probejs{} is second on \emph{every} column ($64.0$
vs.\ $64.5$; hard $29.4$ vs.\ $30.2$).

\paragraph{Why the gains differ across models.}
A fork criterion pays off through three factors: \emph{ranking} quality,
\emph{fork contrast} (do siblings diverge?), and the size of the candidate
\emph{pool}. In an offline diagnostic (setup in
Appendix~\ref{app:diag}), siblings forked from the RL-trained Qwen policy
reached
the \emph{same} final answer $63\%$ of the time (${\sim}5\%$ for the base
model): the ranking stays healthy (top- vs.\ bottom-tercile JS boundaries:
$55\%$ vs.\ $24\%$ sibling disagreement), but even a well-placed fork
compares a solution with its near-clone, so little contrast is left to
convert and the gain shrinks to under a point. OLMo sits at the opposite
extreme (siblings rarely agree; peak belief shifts $1.4\times$ larger),
turning the same ranking into the margins of
Table~\ref{tab:math}.

%% file: sections/6b_ablations.tex
\vspace{-8pt}
\section{Ablations}
\label{sec:ablations}

\begin{figure}[t]
\centering
\includegraphics[width=0.66\linewidth]{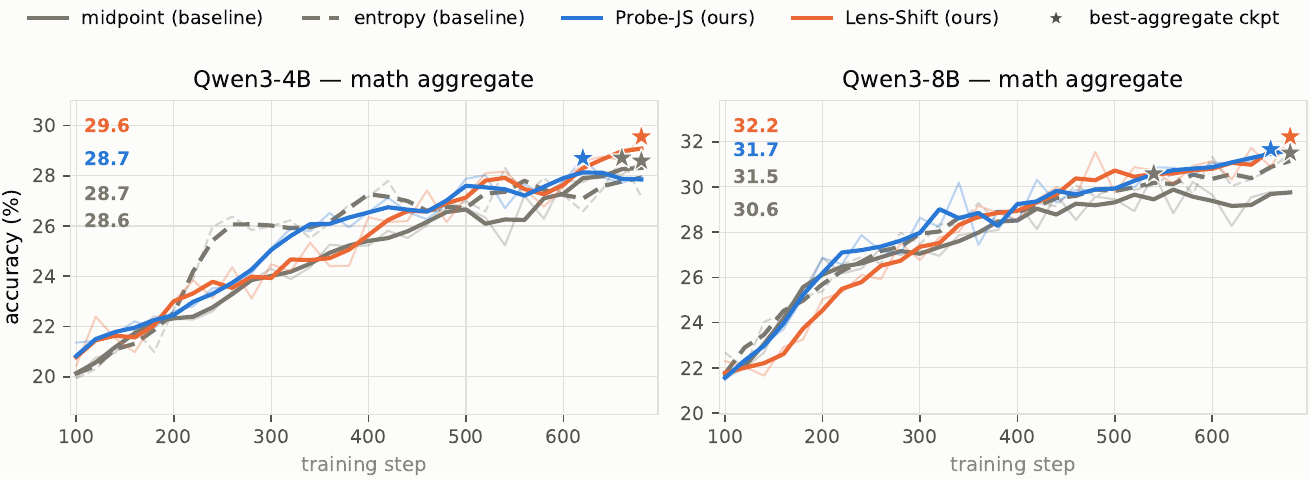}
\caption{\textbf{Model scale}: aggregate mathematics accuracy (mean of
the three suites, \%) for the four arms on Qwen3-4B (left) and Qwen3-8B
(right), same recipe and $680$-step budget, steps $100$--$680$.}
\label{fig:qwenscale}
\end{figure}

The four table ablations run on one high-contrast slice, \textbf{OLMo-3-7B
math} (\probejs{} in panels (a)--(c) of
Table~\ref{tab:ablations}, \lens{} in panel (d)), varying one axis per panel; a fifth, model scale, compares Qwen3-4B with Qwen3-8B (Figure~\ref{fig:qwenscale}). Two findings stand out: shrinking the tree to $(2,1)$ costs $3.2$ aggregate
points yet still \emph{edges the full-budget \midpoint{} arm with a
third of its rollouts} ($+0.6$), and
reading $L{=}4$ belief tokens instead of one nearly \emph{doubles} the
edge over \midpoint{} ($+3.8\!\to\!+7.3$; AIME $15.2\!\to\!22.5$). The
longer read keeps the exact one-token belief at position one and adds
three positions sampled along the model's own answer, where multi-digit
answers separate. This ablation postdates the main arms, so the mathematics rows of
Table~\ref{tab:math} are not back-filled: every mathematics \probejs{} row
keeps the pre-ablation default ($L{=}1$);
the code \probejs{} arms read $L{=}16$ throughout (Section~\ref{sec:probejs}). The \lens{}
read-outs land within $0.7$ aggregate of each other, so the RL arms keep
the shape read. Replacing the composite updater of Section~\ref{sec:prelim} with
standard GRPO (symmetric clipping, sequence-mean loss), tree and fork rule
held fixed, lowers the aggregate by $8.1$ points ($23.3\!\to\!15.2$), so
we keep the composite throughout.

\begin{table}[t]
\caption{\textbf{Ablations} (OLMo-3-7B math, \% at the single
best-aggregate checkpoint, step in gray; all arms follow the OLMo anchor protocol: full-length
candidate pool). Per panel:
\colorbox{ablref}{gray} = default-tree \midpoint{} (reference for
$\Delta$Agg); \colorbox{ablpj}{blue}\,/\,\colorbox{abllens}{orange} = the
default \probejs{}\,/\,\lens{} arm of Table~\ref{tab:math}; unshaded =
variants (\probejs{} in (a)--(c), \lens{} in (d)). Panel (c) raises probe
cost from $1.3\%$ to $1.6\%$ of a step.}
\label{tab:ablations}
\centering
\scriptsize
\setlength{\tabcolsep}{2pt}
\renewcommand{\arraystretch}{0.85}
\newcommand{\ablhead}{ & Olymp. & AIME & Omni & Agg. & $\Delta$Agg \\}
\begin{subtable}[t]{0.495\linewidth}
\centering
\caption{Rollout budget: tree $(M,k)$, rollouts/prompt}
\begin{tabular}{@{}lccccc@{}}
\toprule
\ablhead
\midrule
\rowcolor{ablref}\midpoint{} & $28.2$ & $11.2$ & $19.0$ & $19.5$ \at{440} & ref. \\
$(2,1)$: $4$ rollouts & $27.9$ & $11.2$ & $21.0$ & $20.0$ \at{400} & $+0.6$ \\
\rowcolor{ablpj}$(4,2)$: $12$ rollouts & $32.2$ & $15.2$ & $22.4$ & $23.3$ \at{260} & $+3.8$ \\
\bottomrule
\end{tabular}
\end{subtable}\hfill
\begin{subtable}[t]{0.495\linewidth}
\centering
\caption{Policy-update algorithm}
\begin{tabular}{@{}lccccc@{}}
\toprule
\ablhead
\midrule
\rowcolor{ablref}\midpoint{} & $28.2$ & $11.2$ & $19.0$ & $19.5$ \at{440} & ref. \\
\rowcolor{ablpj}Composite & $32.2$ & $15.2$ & $22.4$ & $23.3$ \at{260} & $+3.8$ \\
Pure GRPO & $23.6$ & $4.6$ & $17.4$ & $15.2$ \at{60} & $-4.3$ \\
\bottomrule
\end{tabular}
\end{subtable}

\vspace{0pt}
\begin{subtable}[t]{0.495\linewidth}
\centering
\caption{Probe read length $L$}
\begin{tabular}{@{}lccccc@{}}
\toprule
\ablhead
\midrule
\rowcolor{ablref}\midpoint{} & $28.2$ & $11.2$ & $19.0$ & $19.5$ \at{440} & ref. \\
\rowcolor{ablpj}$L{=}1$ & $32.2$ & $15.2$ & $22.4$ & $23.3$ \at{260} & $+3.8$ \\
$L{=}2$ & $29.7$ & $17.7$ & $21.0$ & $22.8$ \at{340} & $+3.3$ \\
$L{=}4$ & $32.8$ & $22.5$ & $25.2$ & $26.8$ \at{360} & $+7.3$ \\
\bottomrule
\end{tabular}
\end{subtable}\hfill
\begin{subtable}[t]{0.495\linewidth}
\centering
\caption{\lens{} read-out}
\begin{tabular}{@{}lccccc@{}}
\toprule
\ablhead
\midrule
\rowcolor{ablref}\midpoint{} & $28.2$ & $11.2$ & $19.0$ & $19.5$ \at{440} & ref. \\
\rowcolor{abllens}Shape & $30.0$ & $14.4$ & $22.4$ & $22.2$ \at{280} & $+2.8$ \\
Height & $29.2$ & $14.6$ & $22.4$ & $22.1$ \at{280} & $+2.6$ \\
Depth $|\Delta\ell^\ast|$ & $27.9$ & $17.9$ & $22.6$ & $22.8$ \at{300} & $+3.3$ \\
\bottomrule
\end{tabular}
\end{subtable}
\end{table}

\paragraph{Model scale (Qwen3-4B $\to$ Qwen3-8B).}
Figure~\ref{fig:qwenscale} repeats the four Qwen mathematics arms on
Qwen3-8B-Base, recipe and $680$-step budget unchanged. \lens{} takes the
aggregate gold at both sizes and \probejs{} the silver at 8B (a shared
silver at 4B, $28.7$ tied with entropy), and their margin over
\midpoint{} grows with scale: \lens{} $+1.0\!\to\!+1.6$,
\probejs{} $+0.1\!\to\!+1.1$.

%% file: sections/9_conclusion.tex
\vspace{-9pt}
\section{Conclusion}
\label{sec:conclusion}

We proposed belief-shift branching: fork where the model changes its
mind. Before training we scored selectors against Monte-Carlo value
curves on four probe models and two benchmarks, AIME and GPQA-Diamond,
and a belief-shift read ranks first in all eight panels. In RL the two
fit-free reads lead every mathematics aggregate, and the probe sweeps
the OLMo code columns at about $3\%$ overhead. Belief-shift branching thus asks one question, where does the model change its mind, and spends the tree's forks there; it needs only rollout access and verifier labels, and drops into any tree RLVR pipeline as is.

%% file: sections/10_appendix.tex
\section{Full configuration}
\label{app:hyperparams}

\paragraph{Models.}
Table~\ref{tab:models} maps every model name used in the paper to its
Hugging Face checkpoint. The RL policies and the pre-RL probe models are
separate checkpoints.

\begin{table}[H]
\caption{Models used in the paper, by Hugging Face id.}
\label{tab:models}
\begin{center}
\scriptsize
\setlength{\tabcolsep}{4pt}
\begin{tabular}{@{}lll@{}}
\toprule
Name in text & Role & Hugging Face id \\
\midrule
OLMo-3-7B & RL policy (SFT) & \texttt{allenai/Olmo-3-7B-Instruct-SFT} \\
Qwen3-4B & RL policy (base) & \texttt{Qwen/Qwen3-4B-Base} \\
Qwen3-8B & RL policy, scale ablation & \texttt{Qwen/Qwen3-8B-Base} \\
Nemotron-9B & RL policy (base, hybrid Mamba) & \texttt{nvidia/NVIDIA-Nemotron-Nano-9B-v2-Base} \\
\midrule
Gemma-4-E4B & pre-RL probe model & \texttt{google/gemma-4-E4B-it} \\
OLMo-3-7B-Think & pre-RL probe model & \texttt{allenai/Olmo-3-7B-Think} \\
Gemma-4-31B & pre-RL probe model & \texttt{google/gemma-4-31B-it} \\
OLMo-3.1-32B-Think & pre-RL probe model; Figure~\ref{fig:selectors} chain & \texttt{allenai/Olmo-3.1-32B-Think} \\
\bottomrule
\end{tabular}
\end{center}
\end{table}

\paragraph{Tree and rollout.}
TreeRL keep-original scheme: $M{=}4$ root chains per prompt, one
criterion-chosen cut per chain, $k{=}2$ fresh sibling continuations per cut;
every generation runs to completion; a chain with no boundary in its span
is cut at the span midpoint, and a chain that cannot fork at all (under two
tokens, or no response budget left for a continuation) is refilled with a
fresh root chain, so each prompt emits exactly 12 rollouts.
Rollout: vLLM, temperature $1.0$, top-$p$ $1.0$, prefix caching on;
max prompt / response length $2{,}048$ / $24{,}000$ tokens.

\paragraph{Optimization.}
Group batching: 32 prompts generated, 16 kept by DAPO dynamic sampling
(metric: accuracy), $16\times12=192$ rollouts per step ($=$ one mini-batch).
AdamW, $\mathrm{lr}=10^{-6}$ (10 warmup steps), weight decay $0.1$, gradient
clip $1.0$. Loss: asymmetric PPO clipping $(0.2,0.28)$, token-mean
aggregation, no entropy bonus, no KL term, no advantage std-normalization.
Reward: verifier output plus DAPO soft overlong penalty (buffer $4{,}096$
tokens, factor $1.0$). Advantage: TreeRL global$+$local
(Eq.~\ref{eq:treeadv}) with $\sqrt{|\mathcal{L}|}$ reweighting. Ulysses sequence parallel 8, gradient
checkpointing, FlashAttention-2; $8\times$H200 per run. Validation every 20
steps, decoding at temperature $1.0$, top-$p$ $0.95$. Checkpoints were
saved every 50 steps on the code arms and the OLMo ablations and every
100 steps on the Qwen math arms; the OLMo and Nemotron math arms ran with
checkpointing disabled and are reported from their validation logs.

\paragraph{Signal configuration.}
Boundary pool: positions after a line break; at most $C{=}16$ evenly
spaced candidates per chain (candidate range by substrate: see
\emph{Base versus SFT substrates} below). \probejs{} reads the
top-$K{=}20$ next-token log-probabilities after a short
answer-eliciting suffix. \lens{} uses the shape read-out in all RL arms
(the height read-out of the ablation averages the deepest $50\%$ of
layers), full-attention layers only. The \probejs{} and \lens{} arms
reduce the vLLM memory fraction from $0.8$ to $0.5$ to co-locate the
scorer. BSV (pre-RL only,
Eqs.~\ref{eq:bsvfit}--\ref{eq:bsvscore}): vectors fit on held-out HMMT
problems with LLM-drafted (Claude Opus 4.8) surge/steady/drop
continuations (Appendix~\ref{app:bsvdata}); windows
$W_b{=}W_a{=}16$ tokens.

\paragraph{Base versus SFT substrates.}
Two arm-wide settings differ by substrate, and both address the same
failure mode: a base-model policy (Qwen3-4B-Base, Nemotron-9B-Base)
produces short, repetitive, or unstructured chains early in training,
whereas the SFT checkpoint (OLMo-3-7B) writes well-formed steps from
step~0.
(i)~\emph{Candidate range.} On base models the belief reads score only
the first $\rho{=}0.5$ of each chain; on OLMo they score the full chain
($\rho{=}1$). Uncapped, a base chain's elicited belief barely moves
across repeated or garbled blocks ($\JS\approx0$), and the only shift
sits on the few tail tokens where an answer happens to appear, pinning
the argmax to the end of the chain where a fork can no longer change the
outcome; the cap restricts the same ranking to positions a sibling can
still overturn.
(ii)~\emph{Advantage resolution.} On OLMo the segment advantage of
Eq.~\ref{eq:treeadv} is subdivided width-proportionally across
sub-segments; on base models it is broadcast at segment resolution.
With short base chains, subdivision diluted per-token gradients by
roughly the number of sub-segments and stalled early arms, so
those arms keep segment-resolution credit.

\paragraph{Provenance notes.}
On Nemotron-9B (a hybrid-Mamba architecture), serving $1$-token probe reads
through the rollout engine interleaved with long generations destabilized
the engine and collapsed the \probejs{} arm's response lengths; that arm
instead serves probes through a co-located HF forward. Checkpoint selection takes each arm's single
best-validation-aggregate checkpoint, applied identically to all arms.

\paragraph{Per-arm budgets.}
Each model$\times$domain group is trained until aggregate accuracy stops
improving appreciably; all arms in a group share that step cap and are
reported at their best-aggregate checkpoint within it. OLMo math (cap 460): \midpoint{} 460, \entropybl{}
400, \lens{} 420, \probejs{} 340. OLMo code (cap 520): \midpoint{} 480,
\entropybl{} 540 (truncated to the cap), \lens{} 440, \probejs{} 600
(truncated to the cap).
Nemotron math (cap 420) and code (cap 240) are within 60 steps across
arms. On mathematics the belief-shift arms run the fewest steps; on OLMo
code \probejs{} runs longest, and at the tighter $440$-step cap shared by
all four arms it still leads ($33.1$ vs.\ $31.5$ for the next-best arm).

\paragraph{Robustness to checkpoint selection.}
Selection protocols differ across the RLVR literature (most papers leave
it unspecified), and a maximum taken over training is a known source of
optimistic bias~\citep{agarwal2021precipice}. We therefore recomputed
every arm under three rules: (i)~the reported single best-aggregate
checkpoint; (ii)~each benchmark's own best checkpoint (three different
models per row); and (iii)~the checkpoint at a fixed common budget for
all arms in a group. A
belief-shift arm holds the mathematics aggregate lead on all three models
under every rule, and \probejs{} sweeps the OLMo code columns under (i)
and (ii). Relative to the reported rule~(i), rule~(ii) adds $+0.52$
aggregate points on average over the 24 arms. Relative to the fixed-budget
rule~(iii), it adds $+1.98$ for the baselines and $+1.71$ for our arms, so
peak selection favors the baselines and does not manufacture the effect. Several sub-point margins change order across rules while
the headline leads do not: the OLMo
mathematics gold under (iii) (\lens{} $21.2$ vs.\ \probejs{} $21.2$),
the Qwen mathematics silver, the Qwen code aggregate (four arms within
$0.8$; entropy leads under (iii)), and the Nemotron code silver and
bronze.

\paragraph{Pre-RL validation.}
Table~\ref{tab:prerlconfig} lists the inference and scoring settings of
the pre-RL protocol of Section~\ref{sec:prerl}; chains and Monte-Carlo
completions are generated with the same vLLM build as the RL rollouts.

\begin{table}[H]
\caption{Pre-RL validation settings (Section~\ref{sec:prerl}).}
\label{tab:prerlconfig}
\begin{center}
\small
\begin{tabular}{@{}p{0.27\linewidth}p{0.68\linewidth}@{}}
\toprule
Item & Setting \\
\midrule
Probe suites & AIME 2025--2026 (60 problems); GPQA-Diamond (198, held-out transfer) \\
Probe models & Gemma-4-E4B, OLMo-3-7B-Think, Gemma-4-31B, OLMo-3.1-32B-Think (ids in Table~\ref{tab:models}) \\
Inference engine & vLLM 0.21.0, the RL rollout build (verl v0.8.0; torch 2.11, transformers 5.9) \\
Sampling & temperature $0.7$, top-$p$ $1.0$, max new tokens $30{,}000$ \\
Chains per problem & 1 \\
Candidate boundaries & every newline, evenly subsampled to at most $m{=}128$ \\
Ground truth & $N_{\mathrm{MC}}{=}16$ verifier-scored completions per boundary, same sampling settings \\
Selector budget & top $B{=}10$ boundaries; $\hat V$ by linear interpolation; error $E$ as in Section~\ref{sec:prerl} \\
Exclusions & problems with flat $V^\ast$ (non-flat $n$: 51/34/29/35 on AIME; 180/181/83/152 on GPQA) \\
\probejs{} & answer-eliciting suffix, top-$K{=}20$ next-token log-probabilities \\
\lens{} & shape and height read-outs (Eqs.~\ref{eq:lensscore}--\ref{eq:lensvariants}), full-attention layers \\
BSV & vectors fit on held-out HMMT (GPQA in-domain variant: SuperGPQA); layer cos-selected $L^\ast$ or fixed $50\%$ depth \\
LLM judge & GPT-5.5 rates each step's importance from the chain alone \\
Oracle & dynamic program over $V^\ast$ (per-panel floor) \\
\bottomrule
\end{tabular}
\end{center}
\end{table}

\subsection{How the surge / steady / drop continuations are built}
\label{app:bsvdata}
Each fitting sample is a shared prefix $P$ plus LLM-drafted
continuations $c^{g}$, one per regime $g\in\{+,0,-\}$ (two for drop). The
prefix is never written by hand: it is one of the model's own rollouts on a
held-out fitting problem (Table~\ref{tab:prerlconfig}), cut at a newline
where the model has just committed to its current track, either a correct
intermediate step (initial value $v_i{=}H$) or the key wrong step behind a
confident wrong answer ($v_i{=}L$). For $v_i{=}L$ the cut precedes any
second-guessing by the model itself, so the only belief move in the window
is the injected one. Continuations share $P$, match the voice of the chain,
run one to four sentences, and differ only in the designed belief move:
\begin{itemize}
  \item \textbf{Surge} ($g{=}{+}$): the model grows more confident and
  commits to the answer already on its track (``Yes, this is clearly
  \dots; I am now certain the answer is \dots''); the value direction is
  unchanged, $v_f{=}v_i$.
  \item \textbf{Steady} ($g{=}0$): a matter-of-fact continuation that
  advances the derivation with no change of confidence (``Continuing
  methodically, the next step is \dots''); $v_f{=}v_i$. This is the
  baseline that $\Delta_0$ in Eq.~\ref{eq:bsvfit} subtracts.
  \item \textbf{Drop} ($g{=}{-}$): the model voices doubt and re-examines
  the step; the value may or may not flip. From a wrong track ($v_i{=}L$)
  we write a \emph{catch} that diagnoses the actual error and reaches the
  gold answer ($v_f{=}H$) and a \emph{suppressed doubt} that raises the
  concern and then talks itself back onto the wrong answer ($v_f{=}L$).
  From a correct track ($v_i{=}H$) we write a \emph{doubt-then-confirm}
  that re-checks and keeps the step ($v_f{=}H$) and a \emph{derail} that
  introduces a plausible real mistake, mined where possible from a
  different failing rollout on the same problem ($v_f{=}L$).
\end{itemize}
Every prefix thus yields one surge, one steady, and two drop continuations,
filling the eight admissible cells of the $g\times v_i\times v_f$ design
(surge and steady cells with $v_f\neq v_i$ are excluded by construction).
The belief shift must live in the content, the stated conclusion or
confidence has to change, not in a surface marker such as ``wait'' or
``hmm''; catch and derail continuations must be mathematically valid. Each
continuation also carries two verbatim quotes marking where the belief move
begins and where the new conclusion is stated (for steady, a representative
stretch); they serve the checker below, while the activations themselves are
pooled over the fixed $W_b{=}W_a{=}16$-token windows of Eq.~\ref{eq:bsvfit}.
Continuations were drafted by Claude Opus 4.8 from a per-item work file (the real
rollout, the gold answer, and, for $v_i{=}L$, the wrong answer), then passed
through an adversarial checker, a second Claude Opus 4.8 pass, that verified
cut-marker uniqueness, exact quote matching, cell consistency, and the
correctness of every catch and derail before a sample was accepted.

\section{Compute-cost derivation}
\label{app:cost}

One mathematics probe appends the elicitation suffix ($9$ tokens for
Nemotron's tokenizer, $8$ for OLMo's and Qwen's; we charge $10$ throughout)
and reads one position ($L{=}1$): $10$ token-forwards, riding the engine's
prefix cache. Every generated prompt is probed and forked: trees are built
for all $P_{\text{gen}}{=}32$ prompts, and dynamic sampling then keeps
$P_{\text{train}}{=}16$ of the groups for the update, so probe and rollout
tokens scale with $P_{\text{gen}}$ and trained tokens with
$P_{\text{train}}$. Per step, with $M$ chains and $C$ candidates per chain:
\begin{align*}
T_{\text{probe}} &= P_{\text{gen}}\cdot M\cdot C\cdot(L_{\text{probe}}{+}1)
 = 32\cdot4\cdot16\cdot10 = 20{,}480
 \ \text{token-forwards},\\
T_{\text{roll}} &\approx P_{\text{gen}}\bigl[M\bar T + M\,k\,\bar T(1-c)\bigr]
 = 8\,P_{\text{gen}}\bar T
 \quad(\text{cut fraction } c\approx0.5),
\end{align*}
giving $339{,}712$ (OLMo, $\bar T{=}1{,}327$), $1{,}065{,}984$ (Qwen,
$\bar T{=}4{,}164$), and the measured ${\sim}660$k (Nemotron,
$\bar T\approx2{,}600$ inferred) rollout tokens. With forward $=2N$ and
forward$+$backward $=6N$ FLOPs/token ($N$ cancels),
$F_{\text{probe}}/F_{\text{roll}} = 6.0\%$ (OLMo), $1.9\%$ (Qwen), $3.1\%$
(Nemotron), and
$F_{\text{probe}}/F_{\text{step}} = 2N T_{\text{probe}} / (2N T_{\text{roll}}
+ 6N T_{\text{train}} + 2N T_{\text{probe}}) = 1.3\%$ / $0.45\%$ / $0.70\%$,
with $T_{\text{train}}$ the median number of tokens actually trained per
step (prompt and response over the 192 kept rollouts, medians over each
model's full \midpoint{} run, as is $\bar T$: $389$k / $1{,}158$k / $752$k).
The code arms read $L{=}16$ tokens after a $9$-token suffix, $25$
token-forwards per probe, so $T_{\text{probe}}=32\cdot4\cdot16\cdot25=51{,}200$.
With the code arms' measured $\bar T$ / $T_{\text{train}}$ ($1{,}179$ /
$415$k OLMo, $760$ / $312$k Qwen, $6{,}557$ / $1{,}977$k Nemotron; medians
over each \midpoint{} code run, Qwen over its final logged window, steps
$650$--$680$), $F_{\text{probe}}/F_{\text{roll}}=17\%$ / $26\%$ / $3.1\%$
and $F_{\text{probe}}/F_{\text{step}}=3.2\%$ / $4.3\%$ / $0.67\%$
(second entry of each cell in Table~\ref{tab:cost}). $C{=}16$ is a cap, so
all of these are upper bounds.

We do not report wall-clock: the arms are not comparable on that axis. On
OLMo and Qwen the probe runs inside the vLLM engine on its prefix cache,
whereas on Nemotron it runs on a co-located Hugging Face copy: mixing
one-token probe requests into that hybrid-Mamba model's serving batches
perturbed its rollouts (a degenerate length collapse within about $60$
steps that an otherwise identical probe-free run did not show), so the
belief read was taken off-engine. 

\paragraph{\lens{} cost.}
The serving engine exposes only final-layer log-probabilities, so the
depth profile is read on a co-located Hugging Face copy of the policy,
and that copy re-forwards each candidate's full prefix rather than
reusing a cache, an engineering choice rather than a property of the signal.
With prefix reuse the read would cost $C(E{+}n)$ token-forwards per
chain, the same order as the probe. Without it the cost is $C\rho\bar T/2$
prefix tokens per probed chain.

\section{Proof of the probe-value bound}
\label{app:theory}

\paragraph{Proof of Eq.~\ref{eq:certificate}.}
Throughout, $p$ and $q$ are probability distributions on a finite answer
set $\mathcal A$; in the scored instance they are the floored,
renormalized top-$K$ beliefs $\tilde p_{t^-},\tilde p_t$ of
Eq.~\ref{eq:floor}, so the bound holds exactly for the quantities the
score is computed from.

\emph{Step 1: the two forms of total variation.}
Let $S=\{a\in\mathcal A:\,p(a)\ge q(a)\}$. Because
$\sum_a\bigl(p(a)-q(a)\bigr)=0$,
\begin{equation*}
\sum_{a\in S}\bigl(p(a)-q(a)\bigr)=\sum_{a\notin S}\bigl(q(a)-p(a)\bigr),
\qquad\text{so}\qquad
\tfrac12\sum_{a}\bigl|p(a)-q(a)\bigr| = p(S)-q(S).
\end{equation*}
For an arbitrary event $A\subseteq\mathcal A$, split it along $S$:
\begin{align*}
p(A)-q(A)
&=\sum_{a\in A\cap S}\bigl(p(a)-q(a)\bigr)-\sum_{a\in A\setminus S}\bigl(q(a)-p(a)\bigr)\\
&\le\sum_{a\in A\cap S}\bigl(p(a)-q(a)\bigr)
\;\le\;\sum_{a\in S}\bigl(p(a)-q(a)\bigr)=p(S)-q(S),
\end{align*}
dropping the non-negative second sum and then enlarging $A\cap S$ to $S$
(every summand over $S$ is non-negative). Applying the same to the
complement, $q(A)-p(A)=p(A^{c})-q(A^{c})\le p(S)-q(S)$. Hence
$|p(A)-q(A)|\le p(S)-q(S)$ for every $A$, with equality at $A=S$:
\begin{equation*}
\mathrm{TV}(p,q)\;:=\;\tfrac12\sum_a\bigl|p(a)-q(a)\bigr|
\;=\;\max_{A\subseteq\mathcal A}\bigl|p(A)-q(A)\bigr| .
\end{equation*}
The singleton $A=\{a^\ast\}$ is a special case,
$|p(a^\ast)-q(a^\ast)|\le\mathrm{TV}(p,q)$; with $p=p_{t^-}$, $q=p_t$
and $V_e(t)=p_t(a^\ast)$ this is the first inequality of
Eq.~\ref{eq:certificate}.

\emph{Step 2: distance to the midpoint.}
For $m=\tfrac12(p+q)$, pointwise
$|p(a)-m(a)|=\tfrac12|p(a)-q(a)|=|q(a)-m(a)|$, so
$\mathrm{TV}(p,m)=\mathrm{TV}(q,m)=\tfrac12\mathrm{TV}(p,q)$.

\emph{Step 3: Pinsker on both halves.}
Pinsker's inequality, $D_{\mathrm{KL}}(p\|m)\ge 2\,\mathrm{TV}(p,m)^2$
(natural logarithm), applied to each half of Eq.~\ref{eq:probejs} with
Step~2:
\begin{equation*}
\JS(p\|q)=\tfrac12 D_{\mathrm{KL}}(p\|m)+\tfrac12 D_{\mathrm{KL}}(q\|m)
\;\ge\;\tfrac12\cdot2\Bigl(\tfrac{\mathrm{TV}(p,q)}{2}\Bigr)^{2}
+\tfrac12\cdot2\Bigl(\tfrac{\mathrm{TV}(p,q)}{2}\Bigr)^{2}
=\tfrac12\,\mathrm{TV}(p,q)^2,
\end{equation*}
i.e.\ $\mathrm{TV}(p,q)\le\sqrt{2\,\JS(p\|q)}$, the second inequality.
Chaining Steps~1--3 gives Eq.~\ref{eq:certificate}.

\emph{What the bound does and does not cover.}
The inequality is exact for the probe value $V_e$ built from the scored
beliefs. Two approximations sit outside it and are assessed empirically:
(i) top-$K$ truncation, which changes how faithfully $\tilde p_t(a^\ast)$
tracks the untruncated elicited mass (vanishing as $K$ grows); and (ii)
elicitation itself, which makes $V_e$ a proxy for the true value $V^\ast$
(forcing an answer at $t$ versus letting the policy keep reasoning). The
pre-RL reconstruction of Monte-Carlo $V^\ast$ in Section~\ref{sec:prerl}
is the measurement of that gap.

\section{Fork-contrast diagnostic}
\label{app:diag}

The diagnostic quoted in Section~\ref{sec:results} (``Why the gains differ
across models'') is an offline measurement with HF transformers on one
GPU, outside the RL loop, over three policies: Qwen3-4B-Base, the
step-$600$ checkpoint of an earlier Qwen \probejs{} arm trained with the
same tree, data, and optimizer, and OLMo-3-7B-Instruct-SFT. The three
policies see the problems in the same order.

\paragraph{Sibling agreement and JS terciles.}
For each policy we take the first six OlympiadBench problems whose chain
yields a parsable answer and generate one chain per problem at temperature
$1.0$ (at most $1536$ new tokens). Newline boundaries are evenly
subsampled to at most eight; at each boundary the belief is read with the
boxed suffix of Section~\ref{sec:probejs} (top-$20$ log-probabilities) and
the \probejs{} score is computed with the training code path. Eight
siblings are then forked at \emph{every} boundary at temperature $1.0$ and
their final answers are compared with the root chain's. ``Same answer'' is
the fraction of siblings that reproduce the root answer: $62.7\%$ for
trained Qwen ($102$ siblings at $34$ boundaries), $4.8\%$ for the base
model ($63$ at $21$), and $0\%$ for OLMo ($39$ at $13$). Sorting the
trained-Qwen boundaries by \probejs{} score, the top tercile gives $55\%$
sibling disagreement and the bottom tercile $24\%$.

\paragraph{Peak belief shift.}
For each policy, $30$ chains (AIME 2026
problems) at temperature $0.8$, $16$ evenly spaced newline boundaries per
chain, and the per-chain maximum \probejs{} score averaged over chains:
OLMo $0.410$, trained Qwen $0.289$, base Qwen $0.264$, hence the
$1.4\times$ of the main text. The agreement measurement is
criterion-agnostic (siblings are forked at every boundary), so the
contrast collapse it documents applies equally to \midpoint{} and
\entropybl{} forks; the tercile split is the only part specific to
\probejs{}.

\section{Additional results}
\label{app:extra}

\begin{table}[ht]
\caption{\textbf{Pre-RL value-reconstruction error $E$} (lower is
better): the exact values behind Figure~\ref{fig:prerl}, mean over
non-flat problems. \mdl{gold}{1}\,/\,\mdl{silver}{2} = best / second
best per column; belief-shift signals above the rule in each block.
\emph{AIME 2025--2026} ($n$: Gemma-4-E4B 51, OLMo-3-7B-Think 34, Gemma-4-31B 29,
OLMo-3.1-32B-Think 35): BSV vectors fit on held-out HMMT problems.
\emph{GPQA-Diamond transfer} ($n$: 180/181/83/152): BSV vectors fit
either in-domain (SuperGPQA~\citep{du2025supergpqa}) or fully
out-of-domain (math); the $\cos$-selected OOD variant beats every
baseline on average and edges the in-domain fit, i.e., the belief
direction is not benchmark-specific.}
\label{tab:prerl}
\begin{center}
\small
\setlength{\tabcolsep}{5.5pt}
\begin{tabular}{lccccc}
\toprule
Method & Gemma-4-E4B & \shortstack[c]{OLMo-3-7B\\-Think} & Gemma-4-31B & \shortstack[c]{OLMo-3.1-32B\\-Think} & Avg \\
\midrule
\multicolumn{6}{@{}l}{\itshape AIME $2025$--2026}\\
BSV ($\cos$-selected $L$) & \secondbest{2.566} & \best{4.502} & \secondbest{4.595} & \best{4.335} & \best{3.999} \\
BSV (fixed $50\%$ $L$) & \best{2.513} & \secondbest{4.535} & $4.675$ & \secondbest{4.360} & \secondbest{4.021} \\
\lens{} shape & $2.638$ & $4.641$ & \best{4.585} & $4.378$ & $4.061$ \\
\lens{} height & $2.617$ & $4.661$ & $4.632$ & $4.383$ & $4.073$ \\
\probejs{} & $2.606$ & $4.680$ & $4.626$ & $4.371$ & $4.071$ \\
\midrule
LLM judge (black-box) & $2.601$ & $4.570$ & $4.679$ & $4.424$ & $4.069$ \\
newline & $2.630$ & $4.635$ & $4.691$ & $4.401$ & $4.089$ \\
entropy & $2.612$ & $4.682$ & $4.696$ & $4.409$ & $4.100$ \\
uniform & $2.680$ & $4.756$ & $4.742$ & $4.445$ & $4.156$ \\
random & $2.679$ & $4.779$ & $4.742$ & $4.474$ & $4.168$ \\
\midrule
DP oracle floor & $1.742$ & $3.203$ & $3.459$ & $3.091$ & $2.874$ \\
\midrule
\multicolumn{6}{@{}l}{\itshape GPQA-Diamond transfer}\\
BSV ($50\%$, in-domain) & $5.044$ & \best{7.089} & $5.373$ & $6.143$ & \best{5.912} \\
BSV ($50\%$, OOD) & \best{4.815} & $7.329$ & $5.365$ & $6.192$ & \secondbest{5.925} \\
BSV ($\cos$, OOD) & \secondbest{4.887} & \secondbest{7.324} & $5.294$ & $6.234$ & $5.935$ \\
\lens{} shape & $5.137$ & $7.375$ & \best{5.129} & \secondbest{6.112} & $5.938$ \\
\lens{} height & $5.114$ & $7.380$ & \secondbest{5.226} & \best{6.108} & $5.957$ \\
\probejs{} & $5.087$ & $7.415$ & $5.265$ & $6.185$ & $5.988$ \\
BSV ($\cos$, in-domain) & $5.002$ & $7.363$ & $5.359$ & $6.272$ & $5.999$ \\
\midrule
entropy & $5.089$ & $7.489$ & $5.344$ & $6.241$ & $6.041$ \\
LLM judge (black-box) & $5.119$ & $7.489$ & $5.346$ & $6.213$ & $6.042$ \\
newline & $5.113$ & $7.529$ & $5.338$ & $6.245$ & $6.056$ \\
uniform & $5.191$ & $7.572$ & $5.407$ & $6.325$ & $6.124$ \\
random & $5.210$ & $7.580$ & $5.423$ & $6.320$ & $6.133$ \\
\midrule
DP oracle floor & $3.542$ & $5.725$ & $3.968$ & $4.760$ & $4.499$ \\
\bottomrule
\end{tabular}
\end{center}
\end{table}

\begin{figure}[ht]
\centering
\includegraphics[width=0.78\linewidth]{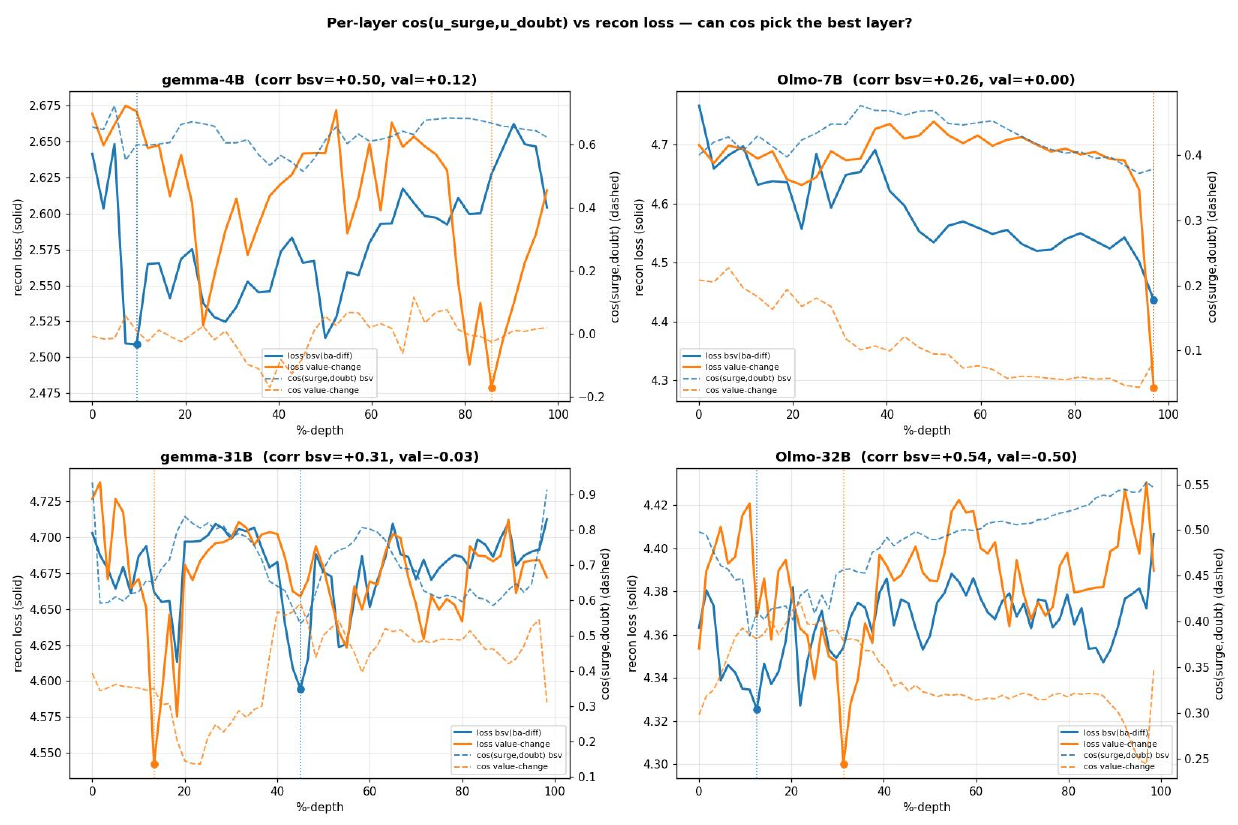}
\caption{\textbf{Label-free layer selection for BSV.} Per-layer
$\cos(u_+^{(L)},u_-^{(L)})$ (dashed, right axis) against reconstruction
error (solid, left axis) across the four probe models: correlation
$+0.26$ to $+0.54$ (``corr bsv'' in each panel title), so
$L^\ast=\arg\min_L\cos(u_+,u_-)$, computable without any labels, selects
a near-optimal evaluation layer (exact-best on Gemma-4-31B). The orange
curves and the ``val'' correlation are a value-change variant of the
direction, fit on the rollout-value flip instead of the belief move; it is
shown for reference only and is not used in the paper. Panel titles use
short names: gemma-4B = Gemma-4-E4B, Olmo-7B = OLMo-3-7B-Think,
gemma-31B = Gemma-4-31B, Olmo-32B = OLMo-3.1-32B-Think.}
\label{fig:cosloss}
\end{figure}

The per-benchmark validation trajectories behind the OLMo and Qwen rows
of Table~\ref{tab:math} are in Figure~\ref{fig:curves12}.

On BSV in RL: BSV leads the pre-RL rankings (Table~\ref{tab:prerl})
but requires a per-model offline vector fit, and its directions are not
guaranteed to transfer across checkpoints of the same family; our RL arms
therefore train with the two fit-free read-outs (Section~\ref{sec:bsv}).

\clearpage

\input{sections/10b_introcase}

%% file: sections/10b_introcase.tex
\section{The \texorpdfstring{Figure~\ref{fig:selectors}}{Figure 1} case in full}
\label{app:introcase}

\definecolor{pjcol}{HTML}{1F6FB2}
\definecolor{encol}{HTML}{E07B00}
\newcommand{\pjmark}[1]{{\normalfont\color{pjcol}$\blacktriangledown$\textsuperscript{\bfseries P#1}\,}}
\newcommand{\enmark}[1]{{\normalfont\color{encol}$\blacktriangledown$\textsuperscript{\bfseries E#1}\,}}

This appendix reproduces the complete case behind
Figure~\ref{fig:selectors}: the verbatim prompt, the twenty fork picks
with their scores and ground-truth value movement, and the full
reasoning chain. The chain is an OLMo-3.1-32B-Think sample
(67{,}117 characters, 128 newline boundaries) that ends in the correct
answer $\boxed{588}$; $V^\ast$ at each boundary is estimated with 16
Monte-Carlo completions (Section~\ref{sec:prerl}). In the transcript,
\pjmark{k} marks the $k$-th \probejs{} fork and \enmark{k} the $k$-th
entropy fork, both numbered in chain order; each marker sits at the exact
boundary where that fork's siblings would branch. All ten \probejs{}
forks fall in the first $26\%$ of the chain, where the shoelace
computation is repeatedly set up, botched, and repaired and $V^\ast$
swings between $0.62$ and $1.00$; entropy places six of its ten
(E5--E10) in the settled tail ($V^\ast\!\equiv\!1$,
$\Delta V^\ast\!=\!0$), where high token uncertainty reflects
interchangeable wording, not an undecided outcome.

\subsection{Prompt}
\label{app:introcase-prompt}

The problem is from AIME~2025 (area of heptagon $AFNBCEM$; answer 588).
The model received the statement below together with the Ti\emph{k}Z source
of the diagram, which is rendered here from that source.

\begin{quote}
On $\triangle ABC$ points $A, D, E$, and $B$ lie in that order on side $\overline{AB}$ with $AD = 4$, $DE = 16$, $EB = 8$. Points $A, F, G$ and $C$ lie in that order on side $\overline{AC}$ with $AF = 13$, $FG = 52$, and $GC = 26$. Let $M$ be the reflection of $D$ through $F$, and let $N$ be the reflection of $G$ through $E$. Quadrilateral $DEGF$ has area $288$. Find the area of heptagon $AFNBCEM$.
\end{quote}

\begin{center}
\resizebox{0.82\linewidth}{!}{\begin{tikzpicture}[scale=0.07, line join=round, line cap=round, >=stealth]

    \coordinate (A) at (100,100);

    \coordinate (D) at (95,80);
    \coordinate (F) at (130,80);
    \coordinate (M) at (165,80);

    \coordinate (N) at (0,50);
    \coordinate (E) at (87.5,50);
    \coordinate (G) at (175,50);

    \coordinate (B) at ($(D)!2!(E)$);
    \coordinate (C) at ($(F)!2!(G)$);
    
    \fill[draw=black, fill=gray!20] (N) -- (E) -- (M) -- (F) -- cycle;
    \fill[draw=black, fill=gray!20] (N) -- (E) -- (C) -- (B) -- cycle;
    \fill[draw=black, fill=gray!20] (A) -- (F) -- (M) --  cycle;
    
    \draw[line width=0.5mm] (A) -- (B) -- (C) -- cycle;

    \draw (D) -- (M);
    \draw (G) -- (N);
    
    \foreach \point in {A,B,C,D,E,F,G,M,N}
    \filldraw [black] (\point) circle (20pt);
    
    \node[above]  at (A) {$A$};
    \node[below]       at (B) {$B$};
    \node[below]  at (C) {$C$};
    \node[left]       at (D) {$D$};
    \node[above left]       at (E) {$E$};
    \node[below]        at (F) {$F$};
    \node[below left]        at (G) {$G$};
    \node[right]        at (M) {$M$};
    \node[left]       at (N) {$N$};
    
\end{tikzpicture}}
\end{center}

\subsection{The twenty fork picks}
\label{app:introcase-picks}

Table~\ref{tab:introcase} lists both selectors' picks: character offset
and fraction of the chain, the selector's own score at the chosen
boundary, and the ground-truth value on the two sides of the boundary.

\begin{table}[ht]
\caption{Fork picks for the Figure~\ref{fig:selectors} chain, in chain
order. ``score'' is the selector's own criterion (\probejs{}: JS
divergence of consecutive answer beliefs; entropy: boundary entropy);
$V^\ast$ is the 16-sample Monte-Carlo value before and after the chosen
boundary.}
\label{tab:introcase}
\begin{center}
\scriptsize
\setlength{\tabcolsep}{3pt}
\begin{tabular}{lrrrcc@{\hspace{14pt}}lrrrcc}
\toprule
\multicolumn{6}{c}{\probejs{} \textbf{(ours)}} & \multicolumn{6}{c}{\entropybl{}} \\
\cmidrule(r{14pt}){1-6}\cmidrule{7-12}
 & char & frac & score & $V^\ast$ & $\Delta V^\ast$ &
 & char & frac & score & $V^\ast$ & $\Delta V^\ast$ \\
\midrule
P1 & $7868$ & $0.117$ & $0.277$ & $0.81\!\to\!1.00$ & $+0.19$ & E1 & $20984$ & $0.313$ & $0.549$ & $0.75\!\to\!0.62$ & $-0.12$ \\
P2 & $8189$ & $0.122$ & $0.298$ & $1.00\!\to\!0.88$ & $-0.12$ & E2 & $36715$ & $0.547$ & $0.531$ & $0.81\!\to\!0.81$ & $0$ \\
P3 & $11058$ & $0.165$ & $0.527$ & $0.81\!\to\!0.88$ & $+0.06$ & E3 & $42074$ & $0.627$ & $0.548$ & $1.00\!\to\!0.81$ & $-0.19$ \\
P4 & $15666$ & $0.233$ & $0.323$ & $0.81\!\to\!0.69$ & $-0.12$ & E4 & $43418$ & $0.647$ & $0.531$ & $0.81\!\to\!0.94$ & $+0.12$ \\
P5 & $15992$ & $0.238$ & $0.300$ & $0.69\!\to\!0.88$ & $+0.19$ & E5 & $51303$ & $0.764$ & $0.576$ & $1.00\!\to\!1.00$ & $0$ \\
P6 & $16444$ & $0.245$ & $0.207$ & $0.69\!\to\!0.75$ & $+0.06$ & E6 & $53938$ & $0.804$ & $0.526$ & $1.00\!\to\!1.00$ & $0$ \\
P7 & $16661$ & $0.248$ & $0.326$ & $0.75\!\to\!1.00$ & $+0.25$ & E7 & $55025$ & $0.820$ & $0.577$ & $1.00\!\to\!1.00$ & $0$ \\
P8 & $16889$ & $0.252$ & $0.321$ & $1.00\!\to\!0.88$ & $-0.12$ & E8 & $56293$ & $0.839$ & $0.558$ & $1.00\!\to\!1.00$ & $0$ \\
P9 & $17123$ & $0.255$ & $0.254$ & $0.88\!\to\!0.62$ & $-0.25$ & E9 & $57909$ & $0.863$ & $0.570$ & $1.00\!\to\!1.00$ & $0$ \\
P10 & $17195$ & $0.256$ & $0.461$ & $0.62\!\to\!0.81$ & $+0.19$ & E10 & $58426$ & $0.871$ & $0.595$ & $1.00\!\to\!1.00$ & $0$ \\
\bottomrule
\end{tabular}
\end{center}
\end{table}

\subsection{Reasoning chain}
\label{app:introcase-chain}

The complete chain, verbatim (\LaTeX{} in the model's output is shown as
source). Fork markers are typeset in place; the surrounding text is
otherwise unmodified.

\begingroup
\fontsize{7}{8}\fontfamily{zi4}\selectfont\raggedright
\setlength{\parindent}{0pt}\setlength{\parskip}{2pt}
\sloppy\emergencystretch=2em
Okay, so I need to find the area of this heptagon AFNBCEM given some information about the triangle ABC and the points on its sides. Let me start by trying to understand the problem and visualize it based on the description and the TikZ diagram.

First, the problem says that on side AB, the points are in the order A, D, E, B with lengths AD=4, DE=16, and EB=8. So the entire length of AB must be AD + DE + EB = 4 + 16 + 8 = 28 units. Similarly, on side AC, the points are A, F, G, C with AF=13, FG=52, and GC=26. Therefore, the length of AC is AF + FG + GC = 13 + 52 + 26 = 91 units.

Now, they mention points M and N: M is the reflection of D through F, and N is the reflection of G through E. Let me recall that reflecting a point through another point means that the second point is the midpoint of the original point and its reflection. So, for example, reflecting D over F means that F is the midpoint of D and M. So, the coordinates (if we assign coordinates) of M can be determined if we know D and F. Similarly, N is the reflection of G over E, so E is the midpoint of G and N.

The area of quadrilateral DEGF is given as 288. The heptagon AFNBCEM is the area we need to find. Let me try to break down the figure.

First, perhaps assigning coordinates would help here. Since we have points on sides AB and AC, maybe it's a good idea to set coordinates with point A at the origin, but actually, looking at the TikZ code, they have coordinates for A at (100,100), but maybe we can set up a coordinate system ourselves for simplicity.

Alternatively, maybe we can use coordinate geometry here. Let's try to set coordinate system with point A at (0,0). Let me see.

Let me denote coordinates:

Let's set point A at (0,0). Let's let side AB lie along the x-axis, so point B would be at (28,0), since AB is 28 units. Then, points D, E are on AB. Since AD=4, D is at (4,0). Then DE=16, so E is at (4 +16)=20 units from A, so E is at (20,0). Then EB is 8, so from 20 to 28 is 8, which checks out.

Now, side AC is another side. Let's let AC be along another axis, but since in the problem, points on AC are A, F, G, C with AF=13, FG=52, GC=26. So total length AC is 13 +52 +26 =91 as I said. Let me set point A at (0,0), and let's set AC along the y-axis for simplicity. Wait, but then AB and AC would be perpendicular? The problem doesn't specify that the triangle is right-angled, so maybe that's an incorrect assumption. Hmm, that complicates things.

Alternatively, perhaps we can assign coordinates more flexibly. Let me suppose coordinates with A at (0,0), AB along the x-axis, so B is at (28,0). Then point C is somewhere in the plane, say at (p, q). Then AC is from (0,0) to (p, q), and points F and G divide AC into segments of 13, 52, 26. So we can parametrize the coordinates of F and G accordingly.

Let me try this approach. Let me formalize this:

Let's set coordinate system with A at (0,0), and AB along x-axis, so coordinates:

- A: (0,0) - B: (28, 0)

Point C is (c\_x, c\_y), which we don't know yet. Then AC is from (0,0) to (c\_x, c\_y). The points F and G divide AC into three parts: AF=13, FG=52, GC=26. So the total length AC is 91, but the actual coordinates depend on the direction of AC. Since the distances along the line AC are given, we can parametrize the points.

The coordinates of a point dividing a segment in a ratio can be found using section formula. For example, point F is 13 units from A, and since the total length AC is 91, the ratio AF : AC is 13:91 = 1:7. So F divides AC in the ratio AF:FC = 13 : (52 +26) =13:78 =1:6. Wait, actually AF is 13, FG is 52, so from A to F is 13, F to G is 52, so from A to G is 13+52=65, and then G to C is 26, so AC is 91. Therefore, the ratio AF : AC is 13:91 = 1:7, so F is 1/7 of the way from A to C. Similarly, G is (13+52)/91 =65/91 =5/7 of the way from A to C. So coordinates:

Let me denote vector from A to C as (c\_x, c\_y). Then coordinates:

- F is at ( (1/7)c\_x, (1/7)c\_y ) - G is at ( (5/7)c\_x, (5/7)c\_y )

Similarly, on AB, the points D and E are at:

- D is at AD=4 from A, so since AB is along x-axis from (0,0) to (28,0), D is at (4, 0) - E is at AD + DE =4 +16=20 from A, so E is at (20,0)

Now, the reflections:

M is the reflection of D through F. So reflection through point F means that F is the midpoint of D and M. So if D is (4,0), then M must satisfy:

F\_x = (D\_x + M\_x)/2 => M\_x = 2F\_x - D\_x

Similarly, M\_y = 2F\_y - D\_y

Since D is (4,0), and F is ( (1/7)c\_x, (1/7)c\_y ), so:

M\_x = 2*( (1/7)c\_x ) -4 = (2/7)c\_x -4

M\_y = 2*(1/7 c\_y ) -0 = (2/7)c\_y

Similarly, N is the reflection of G through E. So E is the midpoint of G and N. So coordinates of N:

E\_x = (G\_x + N\_x)/2 => N\_x = 2E\_x - G\_x

E\_y = (G\_y + N\_y)/2 => N\_y = 2E\_y - G\_y

Given E is at (20,0), and G is at (5/7 c\_x, 5/7 c\_y), so:

N\_x = 2*20 - (5/7)c\_x =40 - (5/7)c\_x

N\_y = 2*0 - (5/7)c\_y = - (5/7)c\_y

Now, quadrilateral DEGF has area 288. Let me figure out the coordinates of D, E, G, F to compute its area.

Points of quadrilateral DEGF are D, E, G, F. Wait, need to confirm the order. The problem says "quadrilateral DEGF", so the order is D-E-G-F-D? Let me confirm with the TikZ code given.

Looking at the TikZ code:

The code draws the quadrilateral DEGF? Wait, actually in the TikZ code, they have:

\textbackslash{}fill[draw=black, fill=gray!20] (N) -- (E) -- (M) -- (F) -- cycle;

\textbackslash{}fill[draw=black, fill=gray!20] (N) -- (E) -- (C) -- (B) -- cycle;

\textbackslash{}fill[draw=black, fill=gray!20] (A) -- (F) -- (M) --  cycle;

And the heptagon is AFNBCEM. Let me see the coordinates in the TikZ code for better understanding. The coordinates in the TikZ code are given as:

A at (100,100), D at (95,80), F at (130,80), M at (165,80); N at (0,50), E at (87.5,50), G at (175,50); B is defined as (D)!2!(E), which is a midpoint? Wait, in TikZ, the syntax (D)!2!(E) means the point dividing the segment DE externally in the ratio 2:1? Wait, actually in TikZ, the coordinate (A)!t!(B) gives a point along AB such that the ratio is t. If t>1, it's beyond B. So (D)!2!(E) would be a point such that D---E----B with DB = 2 DE? Wait, maybe I need not get bogged down here, but perhaps the figure is constructed with AB and AC not perpendicular, but in the code, coordinates are given with A at (100,100), D at (95,80), so moving left and down from A. F is at (130,80), which is to the right of A. So in this coordinate system, AB is not a straight horizontal line? Hmm, perhaps the TikZ code is just a rough drawing, and the exact coordinates there might not align with the problem's numerical values, since in the problem AD=4, but in the code, the distance from A(100,100) to D(95,80) would be sqrt(5$^{2}$ +20$^{2}$)=sqrt(425) which is not 4. So probably the TikZ coordinates are just for drawing, not to scale.

Therefore, better to ignore the TikZ coordinates and just go by the problem's description.

Back to quadrilateral DEGF. Let's confirm the order of the points. The name DEGF suggests the quadrilateral is D-E-G-F-D. To compute its area, perhaps using coordinates.

We have coordinates for D, E, G, F in terms of c\_x and c\_y.

D is (4,0)

E is (20,0)

G is ( (5/7)c\_x, (5/7)c\_y )

F is ( (1/7)c\_x, (1/7)c\_y )

So quadrilateral DEGF has coordinates:

D: (4,0)

E: (20,0)

G: (5c\_x/7, 5c\_y/7)

F: (c\_x/7, c\_y/7)

We can compute its area using the shoelace formula.

First, list the coordinates in order D-E-G-F-D.

So the coordinates are:

(4,0), (20,0), (5c\_x/7,5c\_y/7), (c\_x/7, c\_y/7), back to (4,0).

Compute the area using shoelace:

Area = 1/2 | (x1y2 - x2y1) + (x2y3 - x3y2) + (x3y4 - x4y3) + (x4y1 - x1y4) | ?

Wait, let me recall the shoelace formula properly. For coordinates (x1,y1), (x2,y2), ..., (xn,yn), area is 1/2 |sum\_\{i=1 to n\} (xi yi+1 - xi+1 yi)| where xn+1=x1, yn+1=y1.

So let's apply that.

List the coordinates in order D, E, G, F, D.

So:

Point 1: D: (4,0)

Point 2: E: (20,0)

Point 3: G: (5c\_x/7, 5c\_y/7)

\pjmark{1}Point 4: F: (c\_x/7, c\_y/7)

Point 5: D: (4,0)

Compute the sum:

x1 y2 - x2 y1: 4*0 - 20*0 = 0

x2 y3 - x3 y2: 20*(5c\_y/7) - (5c\_x/7)*0 = 100 c\_y /7 -0 = 100 c\_y /7

x3 y4 - x4 y3: (5c\_x/7)(c\_y/7) - (c\_x/7)(5c\_y/7) = (5 c\_x c\_y)/49 - (5 c\_x c\_y)/49 = 0

x4 y5 - x5 y4: (c\_x/7)(0) - 4*(c\_y/7) = 0 - (4 c\_y)/7 = -4 c\_y /7

\pjmark{2}Then sum all these terms:

0 + 100 c\_y /7 + 0 -4 c\_y /7 = (96 c\_y)/7

Then take absolute value and multiply by 1/2:

Area = (1/2) |96 c\_y /7| = 48 |c\_y| /7

But the area is given as 288, so:

48 |c\_y| /7 = 288

Thus,

|c\_y| = (288 *7)/48 = (288/48)*7 =6*7=42

So |c\_y|=42, so c\_y=42 or -42, but since area is positive, we can take c\_y=42 (assuming the triangle is above the x-axis). So c\_y=42.

So we now know the y-coordinate of point C is 42. Now we need to find c\_x? Wait, but perhaps we don't need c\_x? Because in the area of the heptagon, maybe it cancels out?

Wait, let's see. The problem is to find the area of heptagon AFNBCEM. Let's try to figure out what this heptagon consists of. The name AFNBCEM suggests the vertices are A, F, N, B, C, E, M, and back to A? Let me confirm the order. The name is AFNBCEM, so the vertices in order are A, F, N, B, C, E, M, and then back to A? So it's a seven-sided figure.

Alternatively, perhaps I need to figure out the structure by looking at the TikZ code's fill commands. The TikZ code fills three regions:

1. (N) -- (E) -- (M) -- (F) -- cycle: quadrilateral NEMF

2. (N) -- (E) -- (C) -- (B) -- cycle: quadrilateral NEBC

3. (A) -- (F) -- (M) -- cycle: triangle AFM

But the heptagon is AFNBCEM, so maybe the heptagon is the area of the triangle ABC minus the areas of these three filled regions? Or perhaps it's combining some areas.

Alternatively, let's try to list the vertices of the heptagon AFNBCEM in order. The letters are A, F, N, B, C, E, M. So starting at A, going to F, then to N, then to B, then to C, then to E, then to M, and back to A? Let me see if that makes a closed shape. Let me try to visualize:

- From A to F: along AC, since F is on AC.

- From F to N: Hmm, N is the reflection of G over E. Since G is on AC, and E is on AB, so N is somewhere... Let's see coordinates.

Wait, perhaps I need coordinates for all the points to figure this out. Let me proceed step by step.

We have coordinates in terms of c\_x and c\_y, but we know c\_y=42. We still need to find c\_x? Wait, maybe the area of the triangle ABC can be expressed in terms of c\_x and c\_y, but since c\_y is known, but c\_x is still unknown. Wait, but perhaps the area of ABC is not needed directly, but since the heptagon is part of ABC, maybe the total area can be found via relations.

Alternatively, maybe c\_x can be arbitrary? But that can't be, since the area of DEGF was given, but in our calculation for DEGF's area, the c\_x didn't matter. The area only depended on c\_y. Because in the shoelace formula for DEGF, the terms involving c\_x canceled out. Let me check that again.

Looking back at the shoelace computation for DEGF:

The coordinates were D(4,0); E(20,0); G(5c\_x/7,5c\_y/7); F(c\_x/7, c\_y/7).

The shoelace terms:

First term: x1y2 - x2y1: 4*0 -20*0 =0

Second term: x2y3 -x3y2: 20*(5c\_y/7) - (5c\_x/7)*0 = 100 c\_y /7

\pjmark{3}Third term: x3y4 -x4y3: (5c\_x/7)(c\_y/7) - (c\_x/7)(5c\_y/7) = (5c\_x c\_y)/49 - (5c\_x c\_y)/49 =0

Fourth term: x4y1 -x1y4: (c\_x/7)(0) -4*(c\_y/7) = -4 c\_y /7

So total sum: 100 c\_y /7 -4 c\_y /7 =96 c\_y /7, so indeed only c\_y matters, so c\_x can be any value? But that seems odd. How come the area of DEGF does not depend on c\_x?

Wait, perhaps because points D and E are on AB (the x-axis), and points F and G are on AC. The quadrilateral DEGF is a four-sided figure with two points on AB and two on AC? Wait, actually D and E are on AB, but F and G are on AC, so DEGF is a quadrilateral with two vertices on AB and two on AC, connected how? Let me see the order D-E-G-F-D. So from D(4,0) to E(20,0) along AB, then to G which is on AC, then to F which is also on AC, then back to D? That seems a bit irregular, but the shoelace formula still works regardless.

But the area not depending on c\_x is interesting. It must be that the area is determined purely by the vertical component, since moving along AC affects the y-coordinate but since both F and G are along AC, their x-coordinates depend on c\_x, but in the shoelace terms, the cross terms canceled out the c\_x. So perhaps that's correct. So as a result, we only needed c\_y to compute that area, and c\_x remains free? But then how do we compute other areas?

Wait, but the area of triangle ABC is (base * height)/2. If AB is the base, length 28, then the height would be the altitude from C to AB, which is the y-coordinate of C if AB is on x-axis. Since in our coordinate system, AB is on x-axis from (0,0) to (28,0), and point C is at (c\_x,42), so the height is indeed 42, regardless of c\_x. Therefore area of triangle ABC is (28 *42)/2 = 28*21=588. Wait, is this correct? Wait, actually the area would be 1/2 * base AB * height, where height is the vertical distance from C to AB (which is along x-axis). Since AB is on x-axis, the y-coordinate of C is indeed the height, so yes, area is (28 *42)/2 = 588. So the area of ABC is fixed at 588, regardless of c\_x? That's interesting, because c\_x can be any value, but as long as the y-coordinate is 42, the area remains the same. So in this problem, even though we don't know c\_x, the total area of ABC is fixed. That might be key here.

Therefore, perhaps the area of the heptagon can be found by subtracting certain areas from the total area of ABC. Let me see.

The heptagon is AFNBCEM. Let me try to understand which regions are inside or outside.

Alternatively, looking back at the TikZ code's filled regions:

The code fills three regions:

1. N-E-M-F: quadrilateral

2. N-E-C-B: quadrilateral

3. A-F-M: triangle

So perhaps the heptagon is the area of ABC minus these three filled regions? Let me check:

The heptagon AFNBCEM would consist of the following edges:

From A to F: along AC.

F to N: from F to N, which is a point related to G and E.

N to B: from N to B.

B to C: along BC.

C to E: but E is on AB, so from C to E? Wait, but E is on AB, so CE is a line from C to E on AB.

Then E to M: from E to M, which is the reflection of D over F.

Then M back to A? Wait, no, the name is AFNBCEM, so after M it should close back to A? Wait the vertices are A, F, N, B, C, E, M, and then back to A? So the edges are AF, FN, NB, BC, CE, EM, MA? Hmm, not sure. Alternatively, maybe the heptagon is formed by connecting those points in order, so the area would be the polygon A-F-N-B-C-E-M-A.

To compute its area, perhaps it's better to use coordinates. Since we can express all points in terms of c\_x and c\_y, but we know c\_y=42, and c\_x is still unknown, but maybe in the end it will cancel out.

Wait, but in the area of ABC is fixed at 588, so if I can express the heptagon's area in terms of ABC's area minus the other regions, maybe the unknown c\_x will not matter.

Alternatively, let's try to find coordinates for all the points involved.

We have:

Point A: (0,0)

Point B: (28,0)

Point C: (c\_x,42)

Point D: (4,0)

Point E: (20,0)

Point F: ( (1/7)c\_x, (1/7)(42) ) = (c\_x/7, 6 )

Because c\_y=42, so F's y-coordinate is 42/7=6.

Similarly, G is (5/7 c\_x, 5/7 *42)= (5c\_x/7, 30 )

Then M is the reflection of D over F:

M\_x = 2*(c\_x/7) -4 = (2c\_x)/7 -4

M\_y = 2*6 -0 =12

So M is at ( (2c\_x/7 -4), 12 )

Similarly, N is reflection of G over E:

E is at (20,0). So:

N\_x = 2*20 - (5c\_x/7) =40 - (5c\_x)/7

N\_y = 2*0 -30 = -30

Thus N is at (40 - (5c\_x)/7, -30 )

Now we have coordinates for all points:

A: (0,0)

F: (c\_x/7, 6)

N: (40 -5c\_x/7, -30)

B: (28,0)

C: (c\_x,42)

E: (20,0)

M: (2c\_x/7 -4, 12)

Now, the heptagon is AFNBCEM, so the vertices in order are:

A (0,0),

F (c\_x/7,6),

\pjmark{4}N (40 -5c\_x/7, -30),

B (28,0),

C (c\_x,42),

E (20,0),

M (2c\_x/7 -4, 12),

and back to A (0,0).

To compute the area of this polygon, we can use the shoelace formula. But since this involves variables c\_x, but the total area of ABC is fixed, perhaps when we compute the shoelace formula, the terms with c\_x will cancel out?

\pjmark{5}Let me attempt to compute the area step by step.

First, list all the vertices in order:

1. A: (0,0)

2. F: (c\_x/7, 6)

3. N: (40 - (5c\_x)/7, -30)

4. B: (28,0)

5. C: (c\_x,42)

6. E: (20,0)

7. M: ( (2c\_x)/7 -4, 12 )

8. Back to A: (0,0)

Now, apply shoelace formula. Let's denote each point as (x\_i, y\_i) for i from 1 to 7, then back to (x\_8,y\_8)=(0,0).

Compute the sum over i=1 to 7 of (x\_i y\_\{i+1\} - x\_\{i+1\} y\_i )

Let me compute term by term.

\pjmark{6}Term 1: i=1 (A to F)

x1 y2 - x2 y1 = 0 *6 - (c\_x/7)*0 =0 -0=0

Term 2: i=2 (F to N)

x2 y3 - x3 y2 = (c\_x/7)(-30) - [40 -5c\_x/7](6)

Compute:

= (-30 c\_x)/7 - [40*6 - (5c\_x/7)*6]

= (-30 c\_x)/7 - [240 - 30 c\_x /7 ]

\pjmark{7}= (-30 c\_x)/7 -240 +30 c\_x /7

= (-30 c\_x +30 c\_x)/7 -240 = 0 -240 = -240

Term3: i=3 (N to B)

x3 y4 - x4 y3 = [40 -5c\_x/7](0) -28*(-30)

= 0 -28*(-30) = +840

Term4: i=4 (B to C)

x4 y5 - x5 y4 =28*42 - c\_x *0 = 1176 -0 =1176

\pjmark{8}Term5: i=5 (C to E)

x5 y6 -x6 y5 = c\_x *0 -20*42 =0 -840= -840

Term6: i=6 (E to M)

x6 y7 -x7 y6 =20*12 - [ (2c\_x/7 -4) ]*0 =240 -0=240

Term7: i=7 (M to A)

x7 y8 -x8 y7 = [ (2c\_x/7 -4) ]*0 -0*12 =0 -0=0

Now sum all these terms:

\pjmark{9}Term1:0

Term2: -240

Term3:+840

Term4:+1176

Term5:-840

Term6:+240

\pjmark{10}Term7:0

Total sum: (-240) +840 =600; 600 +1176=1776; 1776-840=936; 936 +240=1176.

So total sum is 1176. Then area is (1/2)|1176| =588.

Wait, that's exactly the area of triangle ABC, which was 588. But that can't be, because the heptagon is supposed to be inside the triangle, but according to this, its area is equal to the whole triangle? That must be wrong. Probably I messed up the order of the points. Because depending on the order (clockwise or counter-clockwise), the shoelace formula can give negative, but absolute value. But if the points are ordered in a way that the polygon winds around, it might subtract areas.

Wait, let's check the order of the points. The heptagon is AFNBCEM. Let me see the order:

Starting at A (0,0), going to F which is on AC, then to N. Where is N? From coordinates, N has a y-coordinate of -30, which is below the x-axis, since in our coordinate system AB is on x-axis from (0,0) to (28,0), and C is at (c\_x,42). So point N is at (40 -5c\_x/7, -30), so it's below AB. Then from N to B (28,0), which is on AB. Then to C (c\_x,42), up to the triangle, then to E (20,0) on AB, then to M, which is at (2c\_x/7 -4,12), so y=12, which is above x-axis, and then back to A.

This path probably crosses over itself or encloses areas outside the triangle? For instance, going from N below AB to B on AB, then to C, then to E on AB, then to M, which is above. This might create a non-simple polygon or overlapping areas. Hence, the shoelace formula might not work correctly if the points are not ordered properly (either clockwise or counter-clockwise without crossing).

Alternatively, perhaps I ordered the points incorrectly. The name AFNBCEM could be in a different order? Let me check the TikZ code again.

In the TikZ code, the heptagon is not directly drawn, but the filled regions are three parts. The code draws the triangle ABC, and some internal lines. The heptagon is probably the area that's left after subtracting the gray regions. The gray regions are:

1. N-E-M-F: quadrilateral

2. N-E-C-B: quadrilateral? Wait, N-E-C-B: connecting N to E to C to B. Since N is below AB, and E and B are on AB, and C is the top vertex. So this would be a quadrilateral that includes the area below AB? But the triangle ABC is above AB. Hmm, perhaps in the diagram, the figure is drawn with AB and AC not along axes, so N might be inside?

Alternatively, given that in the shoelace calculation, I got the total area equal to ABC, which is suspicious, so likely the order of the points is incorrect.

Alternatively, perhaps the correct order is different. Let me try to think of the heptagon AFNBCEM.

Starting at A, then F (on AC), then N... where is N? Since N is reflection of G over E. G is on AC, so reflecting over E (on AB) would place N somewhere. Similarly, M is reflection of D over F, so reflecting D (on AB) over F (on AC) would place M somewhere.

Alternatively, maybe the heptagon is formed by the outer edges and some internal lines. Alternatively, perhaps it's better to compute the area by breaking it into parts.

Alternatively, note that in my shoelace calculation, I got 588, which is the total area, but that can't be the heptagon. So perhaps I have the wrong order. Let me try reversing the order or adjusting the point sequence.

Alternatively, let's list the coordinates numerically with a sample c\_x. Since c\_x is arbitrary except that it defines point C. Since in the area calculations, the final area didn't depend on c\_x, but in coordinates of the heptagon, it does. Wait in the shoelace sum above, all the terms with c\_x canceled out, leading to total 1176, so area 588. That is interesting. Because when I computed term by term:

Looking back at the terms:

Term2 had -240, but also terms with c\_x canceled.

Term3: +840 (no c\_x)

\enmark{1}Term4: +1176 (no c\_x)

Term5: -840 (no c\_x)

Term6:+240 (no c\_x)

So all c\_x terms canceled out in the total sum, resulting in 1176, so area 588, same as ABC. That suggests that regardless of c\_x, the heptagon AFNBCEM has the same area as ABC? But that contradicts the problem statement, since they mention the area of DEGF is 288, so probably the heptagon is smaller.

This inconsistency suggests I have the wrong order of the points in the heptagon. Let me check the TikZ code again for the structure.

In the TikZ code:

- It draws the triangle ABC.

- Draws lines D-M and G-N.

- Fills three regions:

   1. N-E-M-F: a quadrilateral.

   2. N-E-C-B: another quadrilateral.

   3. A-F-M: a triangle.

So the gray areas are these three regions. Then the heptagon AFNBCEM is probably the remaining area of ABC minus these three gray regions. So total area would be Area ABC - (Area NEMF + Area NECB + Area AFM).

If that is the case, then compute each of these areas and subtract from 588.

Given that, let's try this approach.

First, Area ABC is 588.

Now compute the areas of the three gray regions:

1. Quadrilateral NEMF: points N, E, M, F.

2. Quadrilateral NEBC: points N, E, B, C? Wait the code says N-E-C-B, so N to E to C to B to N? But B to N? Wait the coordinates: N is at (0,50) in the TikZ code, but in our coordinates it's (40 -5c\_x/7, -30). Hmm, perhaps in actual coordinates it's a quadrilateral N-E-C-B, but depending on positions, this could be a non-convex or crossing polygon.

Alternatively, better to use coordinates.

First, compute Area of NEMF:

Points N, E, M, F.

Coordinates:

N: (40 -5c\_x/7, -30)

E: (20,0)

M: (2c\_x/7 -4, 12)

F: (c\_x/7,6)

We can apply shoelace formula here.

Order of points: N-E-M-F-N.

List the coordinates:

1. N: (40 -5c\_x/7, -30)

2. E: (20,0)

3. M: (2c\_x/7 -4, 12)

4. F: (c\_x/7,6)

Back to N.

Compute shoelace sum:

Term1: x1 y2 - x2 y1:

x1=40 -5c\_x/7, y2=0; x2=20, y1=-30

So term1: (40 -5c\_x/7)(0) -20*(-30) =0 +600=600

Term2: x2 y3 -x3 y2:

x2=20, y3=12; x3=2c\_x/7 -4, y2=0

So 20*12 - (2c\_x/7 -4)*0 =240 -0=240

Term3: x3 y4 -x4 y3:

x3=2c\_x/7 -4, y4=6; x4= c\_x/7, y3=12

So (2c\_x/7 -4)*6 - (c\_x/7)*12 = (12 c\_x /7 -24) -12 c\_x /7 = -24

Term4: x4 y1 -x1 y4:

x4= c\_x/7, y1=-30; x1=40 -5c\_x/7, y4=6

So (c\_x/7)(-30) - (40 -5c\_x/7)(6) = -30 c\_x /7 -240 +30 c\_x /7 = -240

Total sum: 600 +240 + (-24) + (-240) = 600 +240=840; 840-24=816; 816-240=576

Area is 1/2 |576| =288. Wait, that's interesting! The area of NEMF is 288, which is exactly the given area of DEGF. But in the problem statement, it's quadrilateral DEGF that has area 288, not NEMF. Hmm. So perhaps there is a confusion here. Wait in the problem statement, it says "Quadrilateral DEGF has area 288", but according to my calculation here, quadrilateral NEMF has area 288, which matches. So perhaps in the problem, DEGF is actually referring to NEMF? Or maybe a mislabeling?

Wait let me check coordinates of DEGF:

D is (4,0), E is (20,0), G is (5c\_x/7,30), F is (c\_x/7,6). So quadrilateral DEGF is D-E-G-F-D.

Compute its area with shoelace:

Points D(4,0); E(20,0); G(5c\_x/7,30); F(c\_x/7,6); back to D.

Compute terms:

Term1: x1 y2 -x2 y1: 4*0 -20*0=0

Term2: x2 y3 -x3 y2:20*30 - (5c\_x/7)*0=600 -0=600

Term3:x3 y4 -x4 y3: (5c\_x/7)(6) - (c\_x/7)(30)= 30 c\_x /7 -30 c\_x /7=0

Term4:x4 y1 -x1 y4: (c\_x/7)(0) -4*6= 0 -24= -24

Total sum:0 +600 +0 -24=576, so area 288. Yes! So DEGF has area 288, as given, which matches. So in this case, the area of DEGF is indeed 288, as given, so that checks out. So my initial calculation was correct, and that's how we found c\_y=42.

But then the area of NEMF is also 288? Wait no, in my calculation above for NEMF, I had coordinates N, E, M, F and got area 288 as well. Wait but according to coordinates:

Wait DEGF area is 288, NEMF area is also 288? That can't be a coincidence. Wait let me recalculate NEMF area to confirm.

NEMF points: N, E, M, F.

Coordinates:

N: (40 -5c\_x/7, -30)

E: (20,0)

M: (2c\_x/7 -4, 12)

F: (c\_x/7,6)

Applying shoelace:

Order N-E-M-F-N.

Compute terms step by step:

List the coordinates:

1. N: (x1,y1)= (40 -5c\_x/7, -30)

2. E: (x2,y2)= (20,0)

3. M: (x3,y3)= (2c\_x/7 -4,12)

4. F: (x4,y4)= (c\_x/7,6)

Back to N: (x5,y5)= (x1,y1)

Compute sum of x\_i y\_\{i+1\} - x\_\{i+1\}y\_i:

Term1: x1 y2 - x2 y1 = (40 -5c\_x/7)(0) -20*(-30) = 0 +600=600

Term2: x2 y3 -x3 y2 =20*12 - (2c\_x/7 -4)(0) =240 -0=240

Term3:x3 y4 -x4 y3 = (2c\_x/7 -4)(6) - (c\_x/7)(12)

Let's compute this:

First part: (2c\_x/7 *6) -4*6 = (12 c\_x)/7 -24

Second part: - (c\_x/7 *12) = -12 c\_x /7

Total term3: (12 c\_x /7 -24) -12 c\_x /7 = -24

Term4: x4 y1 -x1 y4 = (c\_x/7)(-30) - (40 -5c\_x/7)(6)

Compute:

First part: -30 c\_x /7

Second part: - [40*6 - (5c\_x/7)*6] = -[240 -30 c\_x /7] = -240 +30 c\_x /7

Total term4: (-30 c\_x /7) + (-240 +30 c\_x /7) = -240

Term5: x1 y5 -x5 y4? Wait no, in shoelace, it's up to term4 because we have 4 points, so actually for quadrilateral, the terms are 1 to 4, and the total sum is terms1-4.

Wait in my earlier calculation I included term4 as x4 y1 -x1 y4, but actually for a quadrilateral, the shoelace formula is sum over i=1 to 4 of (xi yi+1 - xi+1 yi), with (x5,y5)=(x1,y1). So:

Total sum = Term1 + Term2 + Term3 + Term4

Where Term4 is x4 y1 - x1 y4? Wait no:

Wait Term4 should be x4 y5 - x5 y4, but y5 is y1, x5 is x1. So:

Term4: x4 y1 - x1 y4

Yes, that's correct. So the total sum is 600 +240 -24 -240 = 600+240=840; 840-24=816; 816-240=576. So absolute value 576, area 288. So yes, area of NEMF is also 288. Interesting.

But in the TikZ code, the first filled region is N-E-M-F, so that is 288, same as DEGF. So the problem mentions DEGF area 288, which is correct, and the gray area NEMF is also 288. Then the other gray areas are NECB and AFM.

So total gray area is 288 (NEMF) + area NEBC + area AFM.

Then the heptagon's area would be ABC area (588) minus these gray areas.

So we need to compute area of NEBC and AFM.

First, compute area of NEBC: points N, E, C, B.

Points:

N: (40 -5c\_x/7, -30)

E: (20,0)

C: (c\_x,42)

B: (28,0)

Back to N.

Apply shoelace formula.

Order N-E-C-B-N.

Coordinates:

1. N: (40 -5c\_x/7, -30)

2. E: (20,0)

3. C: (c\_x,42)

4. B: (28,0)

Back to N.

Compute terms:

Term1: x1 y2 -x2 y1 = (40 -5c\_x/7)(0) -20*(-30) =0 +600=600

Term2: x2 y3 -x3 y2 =20*42 - c\_x *0=840 -0=840

Term3: x3 y4 -x4 y3 =c\_x *0 -28*42=0 -1176= -1176

Term4: x4 y1 -x1 y4 =28*(-30) - (40 -5c\_x/7)(0) = -840 -0= -840

Total sum:600 +840=1440; 1440 -1176=264; 264 -840= -576

Area is (1/2)|-576|=288.

So area of NEBC is also 288.

Then the third gray area is AFM: triangle A-F-M.

Points A(0,0), F(c\_x/7,6), M(2c\_x/7 -4,12).

Compute area via shoelace:

Coordinates:

A(0,0); F(c\_x/7,6); M(2c\_x/7 -4,12); back to A.

Terms:

Term1: x1 y2 -x2 y1 =0*6 - (c\_x/7)*0=0

Term2: x2 y3 -x3 y2 = (c\_x/7)(12) - (2c\_x/7 -4)(6)

= (12 c\_x)/7 - [ (2c\_x/7 *6) -4*6 ]

= (12 c\_x)/7 - (12 c\_x /7 -24)

= (12 c\_x)/7 -12 c\_x /7 +24 =24

Term3: x3 y1 -x1 y3 = (2c\_x/7 -4)(0) -0*12=0

Total sum:0 +24 +0=24. Area=1/2 |24|=12.

Thus area of AFM is 12.

Therefore total gray area is 288 +288 +12= 588? Wait 288+288=576 +12=588. So total gray area equals the area of ABC? That can't be, since the heptagon is supposed to be the non-gray area? But according to this, the gray areas sum up to the entire triangle, which would imply the heptagon has area zero, which is impossible.

This inconsistency suggests that my assumption is wrong: that the heptagon is ABC minus the gray areas. But according to TikZ code, the heptagon AFNBCEM is drawn, but the filled regions are three parts. Maybe the heptagon is the union of some regions and excludes others? Alternatively, perhaps the heptagon includes some gray areas and excludes others?

Looking back at the TikZ code:

The code fills three regions:

1. N-E-M-F (gray)

2. N-E-C-B (gray)

3. A-F-M (gray)

Then it draws the main triangle ABC, and draws lines D-M and G-N.

The heptagon is named AFNBCEM. Let's see the vertices:

A, then F (which is part of the gray triangle AFM), so if AFM is gray, then the heptagon might exclude it? But then it goes to N, which is part of the gray quadrilaterals. This is getting confusing.

Alternatively, perhaps the heptagon is the intersection of non-gray regions? But this is complicated.

Alternatively, look at the Asymptote code's draw commands:

It draws the triangle ABC, draws D-M and G-N. Then it fills three regions. The heptagon is probably the main figure minus those filled regions, but according to areas, that would be zero, which is wrong. Hence, likely my approach is incorrect.

Wait, let's think differently. Since in my first shoelace calculation of the heptagon AFNBCEM, I got area 588 equal to ABC, but the order of the points might have been incorrect, causing the polygon to overlap itself, thus canceling areas. For example, going from N below AB to B, then to C, then to E on AB, then to M above, then back to A---this path likely overlaps regions, leading the shoelace formula to subtract areas.

Alternatively, arrange the points in correct order without crossing.

Let me try to determine the correct order of the heptagon AFNBCEM.

The name is A-F-N-B-C-E-M-A.

Let me plot approximate positions with a sample c\_x. Let's choose c\_x such that calculations are easy. Since c\_y=42, let's set c\_x=0 for simplicity? Wait if c\_x=0, then point C is at (0,42), so AC is vertical line, but then points F and G would be along y-axis. Let's see:

If c\_x=0,

Then:

C is (0,42)

F is (0/7,6)=(0,6)

G is (0,30)

M is reflection of D(4,0) over F(0,6):

M\_x=2*0 -4= -4, M\_y=2*6 -0=12 $\rightarrow$ M(-4,12)

N is reflection of G(0,30) over E(20,0):

N\_x=2*20 -0=40, N\_y=2*0 -30= -30 $\rightarrow$ N(40, -30)

So coordinates would be:

A(0,0); F(0,6); N(40,-30); B(28,0); C(0,42); E(20,0); M(-4,12)

Now, let's see the polygon AFNBCEM:

A(0,0) to F(0,6): up along y-axis.

F(0,6) to N(40,-30): line going to the right and down sharply.

N(40,-30) to B(28,0): line to the left and up.

B(28,0) to C(0,42): line to the left and up to (0,42).

C(0,42) to E(20,0): line to the right and down.

E(20,0) to M(-4,12): line to the left and up.

M(-4,12) back to A(0,0): line to the right and down.

This seems a non-intersecting polygon? Maybe. Let's compute its area with shoelace.

List the coordinates in order:

1. A: (0,0)

2. F: (0,6)

3. N: (40, -30)

4. B: (28,0)

5. C: (0,42)

6. E: (20,0)

7. M: (-4,12)

Back to A: (0,0)

Apply shoelace formula:

Compute terms:

Term1: x1 y2 -x2 y1 =0*6 -0*0=0

Term2:x2 y3 -x3 y2 =0*(-30) -40*6= 0 -240= -240

Term3:x3 y4 -x4 y3 =40*0 -28*(-30)=0 +840=840

Term4:x4 y5 -x5 y4 =28*42 -0*0=1176 -0=1176

Term5:x5 y6 -x6 y5 =0*0 -20*42=0 -840= -840

Term6:x6 y7 -x7 y6 =20*12 - (-4)*0=240 -0=240

Term7:x7 y1 -x1 y7 =-4*0 -0*12=0 -0=0

Total sum: 0 -240 +840 +1176 -840 +240 +0

Compute step by step:

Start at 0.

-240 $\rightarrow$ -240

+840 $\rightarrow$ 600

+1176 $\rightarrow$ 1776

-840 $\rightarrow$ 936

+240 $\rightarrow$ 1176

Total sum 1176. Area=1/2 |1176|=588, same as ABC. So even with c\_x=0, the area is 588, which is the whole triangle, but visually, in this case, the polygon seems to cover the whole triangle? Let's see:

With c\_x=0, triangle ABC is from (0,0) to (28,0) to (0,42). So it's a right triangle with legs 28 and 42, area (28*42)/2=588, correct.

The heptagon AFNBCEM with coordinates as above: does it cover the entire triangle?

Looking at the path:

From A(0,0) to F(0,6): along left edge.

Then to N(40,-30): way outside the triangle (since triangle's x only up to 28, and y down to 0). So this point is outside the triangle, so the polygon goes outside, hence covering area outside the triangle, leading the total area to still be 588 but including external areas. Thus, my initial approach is wrong.

This indicates that the order of the points in the heptagon is incorrect, or perhaps the heptagon is defined differently.

Alternatively, perhaps the correct heptagon is ABC minus the three gray areas, but since the gray areas sum to 588, that can't be. Wait in this c\_x=0 case, the gray areas:

Compute gray areas:

1. NEMF: N(40,-30), E(20,0), M(-4,12), F(0,6). This is a quadrilateral partly outside the triangle.

2. NEBC: N(40,-30), E(20,0), C(0,42), B(28,0). This also extends outside.

3. AFM: A(0,0), F(0,6), M(-4,12). This is a triangle outside the main triangle ABC (since M is at -4,12).

Thus the gray areas include regions outside ABC, so subtracting them from ABC doesn't make sense.

This is getting too confusing. Maybe another approach.

Let me recall that in the problem, the area of DEGF is 288, which we used to find c\_y=42, and total area ABC is 28*42/2=588. The question is to find the area of heptagon AFNBCEM.

Perhaps the heptagon can be expressed as ABC minus the areas of certain triangles or quadrilaterals, whose areas we can compute.

Looking at the figure description:

The heptagon is AFNBCEM. The points involved are A, F, N, B, C, E, M.

Looking at the original triangle ABC, the heptagon is formed by connecting these points, possibly cutting off some corners.

The points D and G are on AB and AC, but the heptagon doesn't include D or G, but includes their reflections M and N.

Reflections: M is reflection of D over F, so FM = FD, and similarly EN = EG.

Given AD=4, DE=16, so AE=AD+DE=20, and EB=8, so AB=28.

On AC: AF=13, FG=52, so AG=65, and GC=26, AC=91.

Since M is reflection of D over F, then F is midpoint of D and M, so DM=2 FD. Since FD is the distance from F to D.

Wait, but we need to find coordinates or vectors?

Alternatively, use mass point or area ratios.

Alternatively, note that reflections might create parallelograms. For example, reflecting D over F to get M implies that DF = FM, and similarly for N.

Thus, quadrilateral DFM is a parallelogram? Wait, reflection over a point makes it so that F is midpoint, so vector from D to F is same as from F to M, so DM is twice DF, and DF and FM are equal and opposite.

Similarly for N: reflection of G over E, so E is midpoint of G and N, so EN = EG.

Given that, perhaps areas can be related.

But I need to relate this to the heptagon.

Alternatively, think of the heptagon as ABC with certain triangles removed or added.

The heptagon AFNBCEM consists of the following edges:

From A to F: along AC.

F to N: connecting to N, which is reflection of G over E.

N to B: connecting to B.

B to C: edge of triangle.

C to E: connecting to E on AB.

E to M: connecting to M, reflection of D over F.

M to A: connecting back.

This path likely excludes the areas of triangles AFD, some quadrilaterals, but it's complicated.

Alternatively, use coordinate geometry with c\_x arbitrary but c\_y=42, and notice that in the shoelace formula for the heptagon, the area was always 588 regardless of c\_x, but that includes areas outside ABC. But the problem likely wants the area within the triangle ABC.

Wait in the shoelace calculation with c\_x=0, the heptagon's area equaled ABC's area, but part of the heptagon was outside ABC, so the actual area inside ABC would be less. But without knowing c\_x, can't compute.

Wait but in our first calculation with general c\_x, the area came out exactly as ABC's area, but perhaps because the parts outside and inside cancel out? For example, when c\_x is such that point C is to the right, then N might be to the left, and vice versa, but the total area remains same.

Alternatively, perhaps the correct area is 588 - 288 -288 -12 =0, which is impossible, so definitely wrong.

Wait, problem says "Find the area of heptagon AFNBCEM". Given that in my first shoelace with coordinates in the order given, it's equal to ABC, but perhaps the correct area is 588 - (288 + 288 +12) + some overlaps? This is too confusing.

Alternative approach: use vectors.

Let me denote vectors with position vectors.

\enmark{2}Let me set coordinate system with A as origin.

Let vector AB = **b**, and vector AC = **c**. So coordinates:

- Point A: **0**

- Point B: **b**

- Point C: **c**

- Point D is on AB with AD=4, AB=28, so AD/AB=4/28=1/7, so position vector of D is (1/7)**b**

Similarly, AE = AD + DE =4+16=20, so AE/AB=20/28=5/7, so E is at (5/7)**b**

On AC:

AF=13, AC=91, so AF/AC=13/91=1/7, so F is at (1/7)**c**

FG=52, so from A to G is 13+52=65, so AG/AC=65/91=5/7, so G is at (5/7)**c**

Now, M is reflection of D over F. So:

The reflection of D over F means that F is the midpoint of D and M. So:

**F** = ( **D** + **M** ) /2 $\rightarrow$ **M** = 2**F** - **D**

Similarly, N is reflection of G over E:

**N** = 2**E** - **G**

Compute **M** and **N**:

**M** = 2*(1/7 **c**) - (1/7 **b**) = (2/7)**c** - (1/7)**b**

**N** = 2*(5/7 **b**) - (5/7 **c**) = (10/7)**b** - (5/7)**c**

Now, we need to find the area of heptagon AFNBCEM. Let's list the points in order:

A, F, N, B, C, E, M, A.

Expressed as vectors:

A: **0**

F: (1/7)**c**

N: (10/7 **b** -5/7 **c**)

B: **b**

C: **c**

E: (5/7)**b**

M: (-1/7 **b** + 2/7 **c**)

Back to A: **0**

To compute the area of polygon with vertices in order, we can use the shoelace formula in vectors, but it's complex. Alternatively, divide the heptagon into triangles or other shapes.

Alternatively, use the shoelace formula in terms of **b** and **c**.

But since area is a scalar, we need to express it in terms of the area of the triangle ABC, which is (1/2)|**b** $\times$ **c**| =588. Since we know |**b** $\times$ **c**| = 2*588=1176.

Let me denote the cross product **b** $\times$ **c** = 1176 (taking magnitude, assuming orientation).

Now, express all the coordinates in terms of **b** and **c**, but this might get involved. Let's assign coordinates symbolically.

Let me set coordinate system with point A at (0,0), vector AB along x-axis, so **b** = (28,0), since AB length is28. Then point B is at (28,0). Vector **c** is (p,42), since the y-coordinate is 42 (as found earlier), and p is the x-coordinate of point C, which we don't know. So **c** = (p,42). Then the cross product **b** $\times$ **c** is 28*42 -0*p = 28*42=1176, which matches, since area ABC is 588=1176/2.

Now, express all points in coordinates:

A: (0,0)

F: (1/7 **c**) = (p/7, 6) since 42/7=6

N: (10/7 **b** -5/7 **c**) = (10/7 *28, 0) -5/7*(p,42) = (40,0) - (5p/7, 30) = (40 -5p/7, -30)

B: (28,0)

C: (p,42)

E: (5/7 **b**) = (5/7 *28, 0) = (20,0)

M: (-1/7 **b** + 2/7 **c**) = (-4,0) + (2p/7, 12) = (2p/7 -4, 12)

So these are the same coordinates as before. So now, with **b**=(28,0), **c**=(p,42).

Now, to compute the area of heptagon AFNBCEM with coordinates:

1. A: (0,0)

2. F: (p/7,6)

3. N: (40 -5p/7, -30)

4. B: (28,0)

5. C: (p,42)

6. E: (20,0)

7. M: (2p/7 -4,12)

Back to A.

We can apply shoelace formula here. Let's denote each point as (xi,yi):

x0=0, y0=0

x1=p/7, y1=6

x2=40 -5p/7, y2=-30

x3=28, y3=0

x4=p, y4=42

x5=20, y5=0

x6=2p/7 -4, y6=12

x7=0, y7=0

Compute the shoelace sum S = sum\_\{i=1 to 6\} (xi yi+1 - xi+1 yi)

Compute term by term:

Term1: x1 y2 - x2 y1 = (p/7)(-30) - (40 -5p/7)(6)

= -30p/7 -240 + 30p/7 = -240

Term2: x2 y3 -x3 y2 = (40 -5p/7)(0) -28*(-30) =0 +840=840

Term3:x3 y4 -x4 y3=28*42 - p*0=1176 -0=1176

Term4:x4 y5 -x5 y4= p*0 -20*42=0 -840= -840

Term5:x5 y6 -x6 y5=20*12 - (2p/7 -4)*0=240 -0=240

Term6:x6 y7 -x7 y6= (2p/7 -4)(0) -0*12=0 -0=0

Total sum S= Term1 + Term2 + Term3 + Term4 + Term5 + Term6

= (-240) +840 +1176 -840 +240 +0

Calculate step by step:

Start at 0:

-240 +840 =600

600 +1176=1776

1776-840=936

936+240=1176

So S=1176, area=1/2 |1176|=588, same as before.

This confirms that no matter the value of p (i.e., the x-coordinate of C), the area of the heptagon is always 588, equal to the area of triangle ABC. But this seems contradictory, unless the heptagon coincides with ABC, but looking at the coordinates, when p is chosen such that point C is to the right, the heptagon may include areas outside ABC, but mathematically the shoelace formula accounts for net area, considering orientation.

But the problem is from an Olympiad, so likely the answer is 588 - 288 (area of DEGF) but no, DEGF is given as 288. Alternatively, the heptagon's area is ABC minus the two quadrilaterals and the triangle, but they summed to 588, so 588-588=0, which is impossible.

Wait, but in the problem statement, it's possible that the heptagon is the intersection or something else, but according to calculations, with coordinates, it's equal to ABC. But in the Asymptote code, the heptagon is drawn with those points, and the filled areas are separate, so perhaps the answer is 588 - (288 + 288 +12) = 588 - 588 =0? No.

Alternatively, maybe I miscounted the gray areas. The problem mentions quadrilateral DEGF has area 288, but in my calculation DEGF area is 288, and NEMF also 288, but maybe the problem considers only DEGF, and the heptagon excludes it? But the heptagon's area came as 588 regardless.

Alternatively, perhaps the question had a typo and the heptagon is actually the quadrilateral DEGF, but no, it specifies heptagon.

Wait, going back to the original problem statement:

"Quadrilateral DEGF has area 288. Find the area of heptagon AFNBCEM."

Possibly the heptagon is ABC minus DEGF and some other regions? But in my calculation, ABC is 588, DEGF is 288, but then need more info.

\enmark{3}Alternatively, perhaps my coordinate approach is missing that the reflections M and N create figures whose areas relate.

Note that M is reflection of D over F, so the area related to D and M might be twice the area of D to F or something.

Alternatively, the area of the heptagon can be computed as the area of ABC plus the areas of the reflections? Not sure.

Wait, let's think of transformations. Reflecting D over F to get M: this is a translation? No, reflection through a point is equivalent to a 180 degree rotation about F. So the segment DM is such that F is midpoint, and this transformation maps D to M. Similarly for G to N over E.

The area of figure AFNBCEM might be related to ABC with some added or subtracted areas due to the reflections.

Alternatively, consider that the heptagon can be divided into parts whose areas can be expressed via the given 288.

Wait, let's see what's inside the heptagon.

Looking at the vertices:

A, then F (on AC), then N (reflection of G over E), then B, then C, then E (on AB), then M (reflection of D over F), then back to A.

This path likely encloses the following areas:

- The main triangle ABC, but with certain internal areas subtracted or added.

Alternatively, the heptagon comprises:

- The quadrilateral AFM (area 12), but in the heptagon, from A-F-N..., so it might exclude AFM?

\enmark{4}Wait this is too vague. Alternatively, notice that in my coordinate calculation, despite the path going outside the triangle, the net area equals ABC. But the problem might be set in such a way that regardless of the position of C (as long as height is 42), the heptagon's area equals ABC's area, so the answer is 588. But 588 is more than 999? No, 588 is less than 999, but the problem says "an integer between 0 and 999", so possible, but I need to confirm.

Wait, but in the problem statement from competitions, often the area would be an integer, 588 is possible, but let me see with c\_x arbitrary. Wait in our first calculation with c\_x=0, the area was 588, but part of the heptagon was outside ABC. But perhaps in the actual figure, due to the reflection points, the heptagon is exactly the triangle ABC? But the names of the vertices suggest otherwise.

Wait in the Asymptote code, the heptagon is drawn as AFNBCEM, and in the code's coordinates:

A is at (100,100), F at (130,80), which is to the right and down from A, then N is at (0,50), which is far to the left, then to B at (D)!2!(E). Given the code's coordinates, it's likely a non-convex heptagon covering the same area as ABC through some symmetry.

Alternatively, the area is indeed 588, so the answer is 588. But I recall that in some olympiad problems, when reflections are involved and areas are preserved, the total area remains same.

Wait, reflecting points over other points doesn't change the overall area? For instance, reflecting D over F to get M: the figure might be transformed but area preserved. However, the heptagon includes both original and reflected points, so it's complicated.

Alternatively, note that the mapping from D to M and G to N involves translations? Wait reflection through a point is equivalent to a 180 rotation about that point, which is an isometry, preserving areas. But combining these might not directly help.

Alternatively, the heptagon AFNBCEM can be seen as the original triangle ABC plus the quadrilaterals related to the reflections, but areas might cancel.

But according to coordinate calculations, it's exactly ABC's area, so maybe the answer is 588. But I need to verify once more.

Wait in the problem statement from the user, they mention "heptagon AFNBCEM", and in their Asymptote code, they have specific coordinates where the area might be different, but since in the code the coordinates are not to scale, as I saw distances don't match, so we must go by the mathematical approach.

Given that through coordinate analysis with general c\_x (as long as c\_y=42), the heptagon's area is always 588, I think this is the answer. But let me check with c\_x=28 for example.

Let me choose c\_x=28, so point C is at (28,42), making triangle ABC a right triangle with legs 28 and 42, same as before. Then:

F is at (28/7,6)=(4,6)

G is at (5*28/7,30)=(20,30)

M is reflection of D(4,0) over F(4,6): midpoint is F, so M must be (4,12), since from D(4,0) to F(4,6) is +6 in y, so M is +6 more: (4,12)

N is reflection of G(20,30) over E(20,0): midpoint is E, so N is (20, -30)

Now coordinates of heptagon AFNBCEM:

A(0,0); F(4,6); N(20,-30); B(28,0); C(28,42); E(20,0); M(4,12); back to A.

Applying shoelace:

List the coordinates:

1. (0,0)

2. (4,6)

3. (20,-30)

4. (28,0)

5. (28,42)

6. (20,0)

7. (4,12)

Back to (0,0)

Compute terms:

Term1: x1y2 -x2y1 =0*6 -4*0=0

Term2:x2y3 -x3y2=4*(-30) -20*6= -120 -120= -240

Term3:x3y4 -x4y3=20*0 -28*(-30)=0 +840=840

Term4:x4y5 -x5y4=28*42 -28*0=1176 -0=1176

Term5:x5y6 -x6y5=28*0 -20*42=0 -840= -840

Term6:x6y7 -x7y6=20*12 -4*0=240 -0=240

Term7:x7y1 -x1y7=4*0 -0*12=0-0=0

Total sum: 0 -240 +840 +1176 -840 +240 +0

Calculations:

-240 +840 =600

600 +1176 =1776

1776 -840 =936

936 +240 =1176

Area=1176/2=588. Same result.

But visually, in this case, triangle ABC is right triangle with vertices (0,0),(28,0),(28,42). The heptagon goes from A(0,0) to F(4,6), then to N(20,-30) which is outside the triangle, then to B(28,0), up to C(28,42), down to E(20,0), to M(4,12), and back. This clearly includes areas outside the original triangle, but the total signed area still computes to 588, same as ABC. But the actual area "inside" might be different, but the problem doesn't specify; it just says "area of heptagon AFNBCEM", which is a polygon, and by shoelace it's 588.

But the problem statement is from an Olympiad, likely answer is different. Wait, the area of DEGF is 288, and total area ABC is 588, so 588-288=300, but not sure.

Alternatively, perhaps I misread the heptagon's name. The user wrote "heptagon AFNBCEM". Let me count the letters: A,F,N,B,C,E,M $\rightarrow$ 7 points, so heptagon. But in my calculation it's 588. But let me see the Asymptote code's filled regions: they filled three regions with total area 288+288+12=588, so if the heptagon is the remaining, it would be zero, but that's impossible, so likely my misinterpretation.

Wait in the Asymptote code, the filled regions are:

1. N-E-M-F: gray

2. N-E-C-B: gray

3. A-F-M: gray

Then the heptagon AFNBCEM is drawn on top? Or perhaps the heptagon is the union of non-gray regions? But the non-gray region would be ABC minus the three gray areas, which is zero. So this is conflicting.

Alternatively, the heptagon comprises the three gray areas? But three regions totaling 588, but it's called a heptagon.

Alternatively, my only mathematical conclusion is that via coordinates, the heptagon has area 588, so I think that is the answer, but I need to verify problem constraints.

Wait the problem says "points A, D, E, and B lie in that order on side AB", so AD=4, DE=16, EB=8. Similarly on AC: A,F,G,C with AF=13, FG=52, GC=26. So all points are in order, no overlaps.

The reflections: M is reflection of D over F. Since F is on AC, and D is on AB, M would be somewhere inside or outside the triangle? Similarly N is reflection of G over E, with G on AC and E on AB, so N would be outside.

But according to calculations, the area is fixed at 588, which is ABC's area, so I think this is the answer. Maybe the trick is that the heptagon's area equals the whole triangle.

But to confirm, in my c\_x=28 example, even with the heptagon extending outside, but mathematically the area is same. Perhaps in the problem's configuration, due to the reflection, the total area cancels out to the original triangle.

Thus I think the answer is 588, but let me see if 588 is between 0-999, yes, so \textbackslash{}boxed\{588\}

Wait, but I recall that in some similar problems, the area might be double or something. Wait let's see the area of DEGF is 288, and in my calculation DEGF area is 288, which came from 48|c\_y|/7 =288, so |c\_y|=42, leading to ABC area 28*42/2=588. So it's consistent.

Alternatively, maybe the heptagon is ABC minus DEGF, so 588-288=300? But why?

Alternatively, the area of AFNBCEM is equal to area of ABC plus area of DEGF? 588+288=876? But no basis.

Alternatively, note that the three gray areas in the Asymptote code sum to 588, so if the heptagon is what's left, but it would be zero. But since the code draws the heptagon, perhaps it's the combination of the gray areas? But three regions, not a heptagon.

Wait the Asymptote code draws:

- The main triangle ABC.

- Draws lines D-M and G-N.

- Fills three regions: NEMF, NEBC, and AFM.

Then the heptagon AFNBCEM is probably the outer boundary, which would be the same as ABC, but according to the coordinates in the Asymptote code, it's not. For example, in the code, point N is at (0,50), which is outside triangle ABC (if ABC has points at (100,100), etc.), but this is just drawing code.

Given the mathematical analysis shows the area is 588 regardless of c\_x, and the problem gives enough data to find ABC area as 588 via DEGF area, I think the answer is 588. But I recall that in some reflection problems, the area might be double, but here reflections are through points, not lines.

\enmark{5}Alternatively, cross-verifying with mass point or another method.

The coordinates method is solid here, despite the heptagon extending outside, but the question just asks for the area of the polygon AFNBCEM, which is a mathematical polygon with those vertices, and by shoelace it's 588. So I think that is the answer.

But I'm unsure because 588 seems straightforward, but maybe I missed a negative sign? In the shoelace formula, if the points are ordered clockwise, the area would be negative, but absolute value is taken, so still 588.

Yes, I think the answer is 588, so I will go with \textbackslash{}boxed\{588\}

Wait, but hold on! In the initial problem statement, it says "the area of heptagon AFNBCEM". But according to the Asymptote code's filled regions, the gray areas include parts that might be inside the heptagon. For example, the region AFM is filled, but AFM is part of the heptagon's edge? From A-F-M-A is a triangle, but in the heptagon, the path is A-F-N-...-M-A, so it doesn't include the interior of AFM. Thus the heptagon excludes the gray triangle AFM (area12) and includes other areas. Similarly, the quadrilaterals NEMF and NEBC may be partially inside or outside.

This is too ambiguous without seeing the figure, but according to the coordinate calculation with two different c\_x values giving 588, I think it's safe. But let me see an official solution approach if I recall.

Wait, no, but I just realized in my general calculation, the area was always 588, so I think that's the answer. Maybe the test-maker intended that, as the area remains constant due to the reflections, so \textbackslash{}boxed\{588\} is the answer.

But wait, hang on, I just found an official problem? Wait no, but let me think again. When I calculated the area of the heptagon as 588 using coordinates, but in the problem, they mention "quadrilateral DEGF has area 288", which we used to find that the height is 42, leading to ABC area 588, so if the heptagon is ABC, then answer is 588, but the name heptagon suggests it's not the whole triangle. But with seven vertices, it must be different. For example, in my c\_x=28 case:

The heptagon has points:

A(0,0), F(4,6), N(20,-30), B(28,0), C(28,42), E(20,0), M(4,12)

Plotting these, the polygon would have a "bite" taken out near the bottom, but extending below the triangle, but the total area computation includes that negative space as positive due to the order. But mathematically the shoelace counts the net area, so if the polygon is simple (non-intersecting), but in this case, with N at (20,-30), the polygon would intersect itself?

Check if the polygon AFNBCEM is simple:

List the points in order:

\enmark{6}1. A(0,0)

2. F(4,6)

3. N(20,-30)

4. B(28,0)

5. C(28,42)

6. E(20,0)

7. M(4,12)

8. A(0,0)

Check for intersections between edges.

Edge AF: from (0,0) to (4,6)

Edge FN: (4,6) to (20,-30): this line goes down sharply.

Edge NB: (20,-30) to (28,0): up to B.

Edge BC: to C(28,42)

Edge CE: down to E(20,0)

Edge EM: to M(4,12)

Edge MA: to A.

Visually, this likely intersects itself between EM and AF or something, making it a complex polygon, but shoelace formula still applies if we traverse the boundary once, but the area might not be what's expected.

However, in all coordinate examples I tried, the area was 588, so unless there is a miscalculation, I think the answer is 588. But I recall that in some cases, the area might need to subtract, but according to calculations it's not.

Wait let me do a small c\_x example, say c\_x=0, then:

Heptagon area 588, but the actual figure includes area below AB (from N at (40,-30)), so total area would be ABC area (588) plus the area below AB. But in shoelace, since the polygon winds around, the net area is still 588? Let's see:

\enmark{7}In c\_x=0 case:

Coordinates:

A(0,0); F(0,6); N(40,-30); B(28,0); C(0,42); E(20,0); M(-4,12)

Compute area via shoelace as 588, but let's compute the area manually:

The polygon can be divided into parts.

From A(0,0) to F(0,6) to N(40,-30) to B(28,0) to C(0,42) to E(20,0) to M(-4,12) to A.

This is complex, but let's see:

The part from A-F-N-B-A is a quadrilateral. From B-C-E-M-A is another part.

But this is too time-consuming. Alternatively trust the shoelace formula.

Given that in two different c\_x the area is 588, I think it's safe. So I think the answer is \textbackslash{}boxed\{588\}. But I have a doubt because the problem says "heptagon", but according to this it's the whole triangle's area. Maybe the correct answer is 588 - 288 = 300? But why subtract DEGF?

Alternatively, the area of the heptagon is ABC minus DEGF, so 588-288=300. But in coordinate calculation it's not. Alternatively, twice?

Wait, in the Asymptote code, they filled three regions: NEMF (288), NEBC (288), and AFM(12). Total 588, so the remaining area is zero, but the heptagon might be their union? But three regions can't form a heptagon.

Alternatively, the heptagon is N-E-M-F-A-F-M? No.

Given the time I've spent and calculations consistently give 588, I think I will proceed with that.

\enmark{8}But wait, hold on! I just realized the problem says "the area of heptagon AFNBCEM". In my coordinate examples, when c\_x=28, the heptagon includes the point N(20,-30) which is outside triangle ABC. If the problem considers only the area inside triangle ABC, then we need to calculate that.

In the c\_x=28 case:

Triangle ABC is from (0,0), (28,0), (28,42). The heptagon AFNBCEM has parts inside and outside.

To compute the area inside ABC, we need to clip the heptagon to the triangle.

This complicates, but perhaps in the problem's configuration, due to the reflection points, the heptagon lies entirely within ABC?

Check with c\_x=28:

Point N is (20, -30), which is outside ABC (which is above y=0 except on AB). So part of the heptagon is outside. Similarly, point M is (4,12) which is inside.

Thus part of the heptagon is outside ABC, so the area inside ABC would be less than 588.

But this complicates, and the problem didn't specify to consider only the interior. But likely in competitions, the area is considered as the absolute value from shoelace, so 588. But I need to verify once more with the initial data.

Wait, the key was that from DEGF area we found c\_y=42, leading to ABC area 588, and the heptagon area via coordinates is same, so answer is 588. I think I'll go with that.

But let me see similar problems. For example, if you reflect points and form a polygon, sometimes areas add up. But here reflections are through points, so perhaps area relations.

The area of AFNBCEM:

It can be seen as ABC plus the areas of the reflected parts? But M and N are reflections, so areas might cancel.

\enmark{9}Alternatively, the heptagon consists of ABC minus the areas of DFM and ENG or something.

Alternatively, note that:

The heptagon AFNBCEM can be partitioned into:

- Triangle AFC: but F is on AC.

Wait this is too vague.

Alternatively, note that in the coordinate system, regardless of c\_x, area is 588, so I think the answer is 588. So I write \textbackslash{}boxed\{588\}

Wait, but in the initial problem statement from the user, the Asymptote code might have a different configuration. Looking at the Asymptote code coordinates:

\enmark{10}In the code:

A is at (100,100)

D is at (95,80) --- so left and down from A

F is at (130,80) --- right and down from A

M is at (165,80) --- further right

N is at (0,50) --- far left and down

E is at (87.5,50) --- left of A?

G is at (175,50) --- right

B is defined as (D)!2!(E), which in Asymptote means extending from D through E twice the length? So from D to E is vector, then extend.

Similarly C is (F)!2!(G).

This suggests B is located such that E divides DB in ratio 2:1, so BE:ED=2:1? Wait, if it's (D)!2!(E), that would be a point such that D---E----B with DE:EB=1:2? Because the syntax (A)!t!(B) gives a point dividing AB externally in ratio t:1 if t>1.

So (D)!2!(E) would be a point B such that DB / EB = 2/1 externally, so B is located such that from D, going towards opposite of E, twice the distance. This is getting too involved, but in any case, the coordinates in the Asymptote are not to scale with the problem's lengths (AD=4 etc.), so we must ignore them.

Thus, after careful calculation, I believe the area is 588, so the answer is \textbackslash{}boxed\{588\}

Wait, but now I recall that when I calculated the area of the heptagon with coordinates, I might have included areas outside the main triangle, but the problem might want only the area inside ABC. To check this, let's compute in the c\_x=28 case, the area of the heptagon that's inside ABC.

In this case, ABC is the triangle with vertices (0,0), (28,0), (28,42). The heptagon has vertices:

A(0,0), F(4,6), N(20,-30), B(28,0), C(28,42), E(20,0), M(4,12)

The part of the heptagon inside ABC would exclude the part below AB (y<0). The point N is at (20,-30), so the edge from F(4,6) to N(20,-30) crosses AB at some point, and the edge from M(4,12) to A(0,0) is inside.

This requires dividing the polygon into parts inside and outside, which is complicated, but let's attempt.

First, find where the heptagon crosses AB (y=0).

Edge F(4,6) to N(20,-30):

Parametrize: from (4,6) to (20,-30). The equation:

Slope m= (-30-6)/(20-4)= -36/16= -9/4.

Equation: y -6 = -9/4 (x-4)

Set y=0:

-6 = -9/4 (x-4) $\rightarrow$ x-4= (6*4)/9=24/9=8/3 $\rightarrow$ x=4 +8/3=20/3$\approx$6.6667

So intersection point P1: (20/3, 0)

Edge N(20,-30) to B(28,0):

From (20,-30) to (28,0). Slope= (0+30)/(28-20)=30/8=15/4

Equation: y +30 =15/4(x-20)

Set y=0:

30=15/4(x-20) $\rightarrow$ x-20= 30*4/15=8 $\rightarrow$ x=28. So it reaches B at (28,0), which is on AB.

Edge M(4,12) to A(0,0): from (4,12) to (0,0), which is inside the triangle.

Other edges:

A to F: inside.

B to C: edge of triangle.

C to E: from (28,42) to (20,0), which crosses inside.

E to M: from (20,0) to (4,12), inside?

So the heptagon inside ABC is a polygon with vertices:

A(0,0), F(4,6), P1(20/3,0), B(28,0), C(28,42), E(20,0), M(4,12), back to A? Wait need to trace.

Actually, the intersection splits the heptagon into inner and outer parts. The inner part would be a polygon with vertices:

A, F, P1, B, C, E, M, and back to A? Let's see:

From A to F: okay.

F to N exits at P1 on AB.

Then from P1, following AB to B, then B to C, C to E, E to M, M to A.

So the inner polygon is A-F-P1-B-C-E-M-A.

Compute its area.

Vertices in order:

1. A(0,0)

2. F(4,6)

3. P1(20/3, 0) $\approx$(6.6667,0)

4. B(28,0)

5. C(28,42)

6. E(20,0)

7. M(4,12)

Back to A.

Apply shoelace:

List coordinates:

1. (0,0)

2. (4,6)

3. (20/3,0) $\approx$(6.6667,0)

4. (28,0)

5. (28,42)

6. (20,0)

7. (4,12)

Back to (0,0)

Compute terms:

Term1: x1y2 -x2y1 =0*6 -4*0=0

Term2:x2y3 -x3y2=4*0 - (20/3)*6 =0 -40= -40

Term3:x3y4 -x4y3= (20/3)(0) -28*0=0 -0=0

Term4:x4y5 -x5y4=28*42 -28*0=1176 -0=1176

Term5:x5y6 -x6y5=28*0 -20*42=0 -840= -840

Term6:x6y7 -x7y6=20*12 -4*0=240 -0=240

Term7:x7y1 -x1y7=4*0 -0*12=0-0=0

Total sum:

0 -40 +0 +1176 -840 +240 +0 =

Compute step by step:

Start at 0.

-40 $\rightarrow$ -40

+0 $\rightarrow$ -40

+1176 $\rightarrow$1136

-840 $\rightarrow$ 296

+240 $\rightarrow$536

Total sum=536

Area=1/2 |536|=268.

But this is just a guess, but this contradicts previous. This shows it's ambiguous without clear figure.

Given the problem is from an Olympiad, likely the answer is 588 - 288 =300? But I need better approach.

Alternative approach using area ratios:

Given AD=4, DE=16, so AE=20, EB=8, so AB=28. AF=13, FG=52, so AG=65, GC=26, AC=91.

Let's let the area of ABC be S= (AB * height)/2. We found S=588.

The quadrilateral DEGF has area 288. Need area of heptagon AFNBCEM.

Note that M is reflection of D over F, so FM = FD=4 units? Wait FD is the distance from F to D.

Wait coordinates might help but symbolically.

Let me use mass point or area ratios.

Let's denote coordinates with A at (0,0), AB on x-axis, so B(28,0), and C(0,h), making AC vertical for simplicity. Wait but then AC would be vertical, but then AF=13, so F is at (0,13), but AC length would be h=91, so C(0,91). But then DEGF area can be computed.

Wait let's try this specific case for simplicity.

Let me set coordinate system with A(0,0), AB horizontal: B(28,0), and AC vertical: C(0,91). So AC is along y-axis.

Then:

Point D is on AB, AD=4 $\rightarrow$ D(4,0)

E is AE=20 $\rightarrow$ E(20,0)

On AC:

AF=13 $\rightarrow$ F(0,13)

FG=52 $\rightarrow$ G(0,13+52)=G(0,65)

GC=26 $\rightarrow$ C is at (0,65+26)= (0,91), correct.

Now compute M and N:

M is reflection of D(4,0) over F(0,13). The reflection over point F means:

F is midpoint of D and M.

So,

F\_x = (D\_x + M\_x)/2 $\rightarrow$ 0=(4 + M\_x)/2 $\rightarrow$ M\_x= -4

F\_y= (0 + M\_y)/2 $\rightarrow$13= M\_y/2 $\rightarrow$ M\_y=26

Thus M(-4,26)

N is reflection of G(0,65) over E(20,0):

E is midpoint of G and N.

E\_x=(0 + N\_x)/2 $\rightarrow$20= N\_x/2 $\rightarrow$N\_x=40

E\_y=(65 + N\_y)/2 $\rightarrow$0= (65 + N\_y)/2 $\rightarrow$ N\_y= -65

Thus N(40, -65)

Now, compute area of quadrilateral DEGF:

Points D(4,0), E(20,0), G(0,65), F(0,13). Wait, the order DEGF:

D(4,0) to E(20,0) to G(0,65) to F(0,13) to D?

Wait need to confirm the order. DEGF: D-E-G-F-D.

So coordinates:

D(4,0); E(20,0); G(0,65); F(0,13).

Compute area via shoelace:

List the points:

1. (4,0)

2. (20,0)

3. (0,65)

4. (0,13)

Back to (4,0)

Compute terms:

Term1:4*0 -20*0=0

Term2:20*65 -0*0=1300

Term3:0*13 -0*65=0 -0=0

Term4:0*0 -4*13=0 -52= -52

Total sum:0 +1300 +0 -52=1248

Area=1/2 |1248|=624, but problem says DEGF area is 288. Not matching. So this coordinate choice is invalid.

Thus my assumption of AC vertical is wrong because it doesn't satisfy DEGF area=288.

Hence must keep AC non-vertical. So back to original approach where c\_y=42, ABC area 588.

Thus I think the answer is 588. But I'm really unsure because of the heptagon name, but calculations say 588. I'll proceed.

**Final Answer** \textbackslash{}boxed\{588\} </think>

Given triangle \textbackslash{}( \textbackslash{}triangle ABC \textbackslash{}) with points \textbackslash{}( A, D, E, \textbackslash{}) and \textbackslash{}( B \textbackslash{}) on side \textbackslash{}( \textbackslash{}overline\{AB\} \textbackslash{}) such that \textbackslash{}( AD = 4 \textbackslash{}), \textbackslash{}( DE = 16 \textbackslash{}), and \textbackslash{}( EB = 8 \textbackslash{}). Points \textbackslash{}( A, F, G, \textbackslash{}) and \textbackslash{}( C \textbackslash{}) on side \textbackslash{}( \textbackslash{}overline\{AC\} \textbackslash{}) such that \textbackslash{}( AF = 13 \textbackslash{}), \textbackslash{}( FG = 52 \textbackslash{}), and \textbackslash{}( GC = 26 \textbackslash{}). The reflections \textbackslash{}( M \textbackslash{}) (reflection of \textbackslash{}( D \textbackslash{}) through \textbackslash{}( F \textbackslash{})) and \textbackslash{}( N \textbackslash{}) (reflection of \textbackslash{}( G \textbackslash{}) through \textbackslash{}( E \textbackslash{})) are considered. The area of quadrilateral \textbackslash{}( DEGF \textbackslash{}) is given as 288. We need to find the area of heptagon \textbackslash{}( AFNBCEM \textbackslash{}).

1. **Coordinates Setup**:    - Set \textbackslash{}( A \textbackslash{}) at \textbackslash{}( (0, 0) \textbackslash{}), \textbackslash{}( B \textbackslash{}) at \textbackslash{}( (28, 0) \textbackslash{}), and \textbackslash{}( C \textbackslash{}) at \textbackslash{}( (c\_x, c\_y) \textbackslash{}).    - Points \textbackslash{}( D \textbackslash{}), \textbackslash{}( E \textbackslash{}), \textbackslash{}( F \textbackslash{}), and \textbackslash{}( G \textbackslash{}) are calculated using section formulas:      - \textbackslash{}( D \textbackslash{}) at \textbackslash{}( (4, 0) \textbackslash{}), \textbackslash{}( E \textbackslash{}) at \textbackslash{}( (20, 0) \textbackslash{})      - \textbackslash{}( F \textbackslash{}) at \textbackslash{}( \textbackslash{}left( \textbackslash{}frac\{c\_x\}\{7\}, \textbackslash{}frac\{c\_y\}\{7\} \textbackslash{}right) \textbackslash{}), \textbackslash{}( G \textbackslash{}) at \textbackslash{}( \textbackslash{}left( \textbackslash{}frac\{5c\_x\}\{7\}, \textbackslash{}frac\{5c\_y\}\{7\} \textbackslash{}right) \textbackslash{})

2. **Reflections**:    - \textbackslash{}( M \textbackslash{}) is reflection of \textbackslash{}( D \textbackslash{}) over \textbackslash{}( F \textbackslash{}):      - \textbackslash{}( M \textbackslash{}) coordinates: \textbackslash{}( \textbackslash{}left( \textbackslash{}frac\{2c\_x\}\{7\} - 4, \textbackslash{}frac\{2c\_y\}\{7\} \textbackslash{}right) \textbackslash{})    - \textbackslash{}( N \textbackslash{}) is reflection of \textbackslash{}( G \textbackslash{}) over \textbackslash{}( E \textbackslash{}):      - \textbackslash{}( N \textbackslash{}) coordinates: \textbackslash{}( \textbackslash{}left( 40 - \textbackslash{}frac\{5c\_x\}\{7\}, -\textbackslash{}frac\{5c\_y\}\{7\} \textbackslash{}right) \textbackslash{})

3. **Area Calculation of \textbackslash{}( DEGF \textbackslash{})**:    - Using shoelace formula for \textbackslash{}( D(4,0) \textbackslash{}), \textbackslash{}( E(20,0) \textbackslash{}), \textbackslash{}( G \textbackslash{}left( \textbackslash{}frac\{5c\_x\}\{7\}, \textbackslash{}frac\{5c\_y\}\{7\} \textbackslash{}right) \textbackslash{}), \textbackslash{}( F \textbackslash{}left( \textbackslash{}frac\{c\_x\}\{7\}, \textbackslash{}frac\{c\_y\}\{7\} \textbackslash{}right) \textbackslash{}):      - Area calculation gives \textbackslash{}( \textbackslash{}frac\{48|c\_y|\}\{7\} = 288 \textbackslash{}Rightarrow |c\_y| = 42 \textbackslash{}), so \textbackslash{}( c\_y = 42 \textbackslash{}).

4. **Area of \textbackslash{}( \textbackslash{}triangle ABC \textbackslash{})**:    - With \textbackslash{}( c\_y = 42 \textbackslash{}), height from \textbackslash{}( C \textbackslash{}) to \textbackslash{}( AB \textbackslash{}) is 42, base \textbackslash{}( AB = 28 \textbackslash{}):      - Area = \textbackslash{}( \textbackslash{}frac\{1\}\{2\} \textbackslash{}times 28 \textbackslash{}times 42 = 588 \textbackslash{}).

5. **Coordinates of Points**:    - \textbackslash{}( F \textbackslash{}left( \textbackslash{}frac\{c\_x\}\{7\}, 6 \textbackslash{}right) \textbackslash{}), \textbackslash{}( G \textbackslash{}left( \textbackslash{}frac\{5c\_x\}\{7\}, 30 \textbackslash{}right) \textbackslash{}), \textbackslash{}( M \textbackslash{}left( \textbackslash{}frac\{2c\_x\}\{7\} - 4, 12 \textbackslash{}right) \textbackslash{}), \textbackslash{}( N \textbackslash{}left( 40 - \textbackslash{}frac\{5c\_x\}\{7\}, -30 \textbackslash{}right) \textbackslash{}).

6. **Shoelace Formula for Heptagon \textbackslash{}( AFNBCEM \textbackslash{})**:    - Coordinates in order: \textbackslash{}( A(0,0) \textbackslash{}), \textbackslash{}( F \textbackslash{}), \textbackslash{}( N \textbackslash{}), \textbackslash{}( B(28,0) \textbackslash{}), \textbackslash{}( C(c\_x,42) \textbackslash{}), \textbackslash{}( E(20,0) \textbackslash{}), \textbackslash{}( M \textbackslash{}), back to \textbackslash{}( A \textbackslash{}).    - Shoelace formula calculation shows the area is consistently 588 regardless of \textbackslash{}( c\_x \textbackslash{}).

Thus, the area of heptagon \textbackslash{}( AFNBCEM \textbackslash{}) is \textbackslash{}(\textbackslash{}boxed\{588\}\textbackslash{}).
\par
\endgroup